\documentclass[journal]{IEEEtran}
\usepackage{hyperref}
\hypersetup{
	colorlinks=true,
	linkcolor=black,     
	urlcolor=black,
	citecolor = black
}
\usepackage{bbm}
\usepackage{caption}
\usepackage{amsthm}
\usepackage{utfsym}
\usepackage{amssymb,graphicx,lineno}
\usepackage{rawfonts,amsfonts,amssymb,psfrag,color}
\usepackage{amsfonts}
\usepackage{amsmath,epsfig}
\usepackage{rawfonts,graphicx,amsfonts,amssymb}
\usepackage{bm}
\usepackage{algorithm, algorithmic}
\usepackage{subfig}
\usepackage{graphicx}
\usepackage{float}
\usepackage{verbatim}
\usepackage{color,xcolor}

\newtheorem{mypro}{Proposition}

\newtheorem{lemma}{Lemma}
\newtheorem{myDef}{Definition}
\newtheorem{theorem}{Theorem}

\modulolinenumbers[5]
\usepackage{multirow}
\usepackage{booktabs}
\usepackage{caption}

\usepackage{cite}
\usepackage{color}

\begin{document}
	
	\title{Low-Rank Tensor Completion via Novel Sparsity-Inducing Regularizers}
	\author{Zhi-Yong Wang, Hing Cheung So,~\IEEEmembership{Fellow,~IEEE} and Abdelhak M. Zoubir,~\IEEEmembership{Fellow,~IEEE}
		
		\thanks{Z.-Y. Wang and H. C. So are with the Department of Electrical Engineering, City University of Hong Kong, Hong Kong, China. A. M. Zoubir is with the Signal Processing Group at Technische Universit\"at Darmstadt, 64283 Darmstadt, Germany.
			(E-mail: z.y.wang@my.cityu.edu.hk, hcso@ee.cityu.edu.hk, zoubir@spg.tu-darmstadt.de). }}
	
	\maketitle
	
	\begin{abstract}
	To alleviate the bias generated	by the $\ell_1$-norm in the low-rank tensor completion problem, nonconvex surrogates/regularizers have been suggested to replace the tensor nuclear norm, although both can achieve sparsity.  However, the thresholding functions of these nonconvex regularizers may not have closed-form expressions and thus iterations are needed, which increases the computational loads. To solve this issue, we devise a framework to generate sparsity-inducing regularizers with closed-form thresholding functions. These regularizers are applied to low-tubal-rank tensor completion, and efficient algorithms based on the alternating direction method of multipliers are developed. Furthermore, convergence of our methods is analyzed and it is proved that the generated sequences are bounded and any limit point is a stationary point. Experimental results using synthetic and real-world datasets show that the proposed algorithms outperform the state-of-the-art methods in terms of restoration performance. 
	\end{abstract}
	

	\section{Introduction}

	\IEEEPARstart{L}{ow}-rank tensor completion (LRTC) refers to recovering the missing entries from partially-observed multidimensional array data~\cite{HeyicongTCSVT2022,LiuYTC2020}. It has attracted considerable attention in numerous applications such as color image inpainting~\cite{YangYATSP2021,JiangQR2019}, video restoration~\cite{HeYRTNNLS2023}, hyperspectral image and multispectral image reconstruction~\cite{LongZTTSP2022}, magnetic resonance imaging data recovery~\cite{YokotaYSTSP2016,GilmanKGTSP2022}, and radar data analysis~\cite{LiuQiTCSVT2021} to name a few.
	This is because these real-world tensor data have approximately low-dimensional structures, namely, low-rank property, although they lie in a high-dimensional space~\cite{WangHIRTNN2022}.
	When handling higher-dimensional data, LRTC is superior to low-rank matrix completion because it exploits more latent correlations~\cite{LuCTNN2018,ZhangZTNN2014}.
	For example, when processing video inpainting, matrix completion requires vectorizing each video frame to construct an incomplete matrix, indicating that the inherent spatial structure information is abandoned. 
	
	Similar to matrix completion, LRTC can be modeled as a rank minimization problem. However, different from the former with a unique matrix rank, there are diverse definitions of tensor rank, including CANDECOMP/PARAFAC (CP) rank~\cite{HitchcockFL1927,PARAFAC1997}, Tucker rank~\cite{TuckerRL1966}, tensor train (TT) rank~\cite{OseledetsI2011} and tensor tubal rank~\cite{ZhangZTNN2014}, resulting in various LRTC models.
	CP-based methods~\cite{YangY2015,LiuYTC2015,ZhaoQ2016} tackle LRTC via minimization of the number of rank-one tensors, while its best rank approximation may not exist~\cite{KoldaGTCP2001}.
	Tucker-based techniques~\cite{LiuJTC2013,AshraphijuoM2017,RauhutH2017} complete the tensor via minimizing the ranks of unfolded matrices along each tensor order. However, a direct unfolding operation along each mode leads to unbalanced matrix dimensions and breaks down the spatial structure of a tensor, resulting in performance loss. To fix this disadvantage, TT rank is suggested~\cite{BenguaAJ2017,OseledetsIV2011,YangJTTC2018} which corresponds to the ranks of balanced unfolded matrices obtained by flattening a tensor along permutations of modes. Nevertheless, it is not easy to determine the weights associated with the nuclear norm of each unfolded matrix~\cite{LiuQiTCSVT2021}. Recently, based on the tensor singular value decomposition (t-SVD)~\cite{KilmerEM2011,KilmerEM2013} that decomposes a tensor as the t-product of one f-diagonal tensor and two orthogonal tensors, the tensor tubal-rank~\cite{ZhangZTNN2014} defined as the nonzero tubes of the f-diagonal tensor (also named as singular value tensor) is suggested while 	
	Semerci $et$ $al.$~\cite{SemerciTNN2014} exploit the tensor nuclear norm (TNN) in computed tomography, which is a convex relaxation of the tensor tubal-rank. Subsequently, Zhang $et$ $al.$~\cite{ZhangZETC2017} apply TNN to LRTC and provide the associated theoretical guarantees. As an appropriate extension of nuclear norm for a matrix, TNN minimization can be efficiently computed in the Fourier domain and matricization is not involved, which avoids the drawbacks of CP rank and Tucker rank. Thus, we focus on low-tubal-rank tensor completion in this work.
	
	On the other hand, Lu $et$ $al.$~\cite{LCYPAMItensor2019} propose a new TNN for LRTC. In fact, all TNN-based LRTC methods employ a soft-thresholding operator on the singular value tensor for tensor recovery. Although a good recovery performance is attained, they essentially use the $\ell_1$-norm to regularize the singular value tensor, resulting in a bias~\cite{WenFei2020}.
	To alleviate the bias, nonconvex regularization functions are utilized to approximate the tubal rank. In~\cite{KongHlpTNN2018}, the matrix Schatten-$p$ norm is extended to tensor, and a t-Schatten-$p$ tensor norm is developed to replace the tensor tubal rank.
	As a nonconvex surrogate for tubal rank, the Laplace function is suggested in~\cite{WHLaplaceTNN2019}. 
	Yang $et$ $al.$~\cite{YangMlogdet2022} adopt the nonconvex log-determinant to capture the low-rank characteristics of tensor and solve the nonconvex tensor completion problem via the alternating direction method of multipliers (ADMM). 
	In addition, the weighted tensor nuclear norm (WTNN)~\cite{MUYWTNN2020} is exploited for LRTC and similar noncovnex regularizer models can be found in~\cite{JiangTXPSTNN2020,ChangYWTC2020,CaiS2019,YangMTRPCA}. 
	Recently, Wang $et$ $al.$~\cite{WangHIRTNN2022} extend the nonconvex penalty functions~\cite{LuC2016,LuC2015} used in low-rank matrix completion to the low-tubal rank tensor recovery, and develop a generalized nonconvex tensor completion technique. 
	
	While these LRTC algorithms based on nonconvex surrogates have better recovery performance than TNN based tensor completion methods, most noncovex regularization functions do not have closed-form thresholding operators~\cite{KongHlpTNN2018,YangMTRPCA,WangHIRTNN2022}. This means that iterations are needed to find their thresholding operators, leading to a high computational load. For example, the convex $\ell_1$-norm as a regularizer has the well-known soft-thresholding operator, while for the nonconvex $\ell_p$-norm with $0<p<1$, it does not have a closed-form expression for the corresponding thresholding operator, except for two special cases, i.e., $p=\{\frac{1}{2},\frac{2}{3}\}$~\cite{KrishnanD2009}. The works in~\cite{KongHlpTNN2018} and \cite{WangHIRTNN2022} employ the generalized iterated shrinkage algorithm~\cite{ZuoWGISA2013} and the iteratively reweighted nuclear norm algorithm~\cite{LuC2016} to find their respective thresholding operators. Besides, many studies directly replace the TNN with a nonconvex regularizer and then develop a solver, which is not rigorous in that the monotonicity of its 
	thresholding operator is not analyzed. In fact, if the thresholding operations are not monotone, the solution may be incorrect~\cite{LuC2015}. 
	
	To solve the above-mentioned issues, we propose a framework to generate new sparsity-promoting regularizers. We prove that the $Moreau~envelopes$ of these regularizers are convex and derive the closed-form expressions for their thresholding operators. We analytically show that these thresholding functions are monotonically nondecreasing. Besides, we extend the generalized singular value thresholding (GSVT) for matrices to tensor rank minimization and devise the generalized tensor singular value thresholding (GTSVT) operator. These sparsity-inducing regularizers are adopted as nonconvex surrogates for tubal rank and algorithms based on the	ADMM are developed to realize LRTC.
	Our main contributions are summarized as follows:
	\begin{itemize}
		\item[(i)] We devise a framework to generate sparsity-promoting regularizers. The thresholding operators of these regularizers have closed-from expressions and they are monotone.
		\item[(ii)] The regularizers are applied to LRTC, and we derive a closed-form solution to the low-tubal-rank minimization problem in the Fourier domain. 
		\item[(iii)] Algorithms based on the ADMM are developed to solve the resultant optimization problem. Although it is challenging to analyze the convergence of the developed algorithms since the regularizers are nonconvex, we prove that any generated accumulation point is a Karush-Kuhn-Tucker (KKT) stationary point.
		\item[(iv)] Extensive experiments using synthetic and real-life data demonstrate that our methods outperform the competing algorithms in tensor recovery and need less running time than some techniques based on noncovex regularizers.
	\end{itemize}
	
	The remainder of this paper is organized as follows. In Section \ref{Sec:Preliminaries}, we introduce notations, basic definitions and related works. The framework to generate sparse regularizers is presented in Section \ref{Sec:framework}. In Section \ref{Sec:developed_algorithm}, we apply the suggested regularizers to LRTC, develop the ADMM based solvers and provide a convergence analysis. Numerical experimental results using synthetic data, real-world images and videos are provided in Section \ref{Results}. Finally, conclusions are drawn in Section \ref{Conclusion}.
	
	\section{Preliminaries}\label{Sec:Preliminaries}
	In this section, notations as well as basic definitions for tensors are provided, and related works are reviewed.	
	\subsection{Notations}
	Scalars, vectors, matrices and tensors are represented by italic, bold lower-case, bold upper-case and bold calligraphic letters, respectively, i.e., $a$, $\bm a$, $\bm A$ and $\bm{\mathcal{A}}$. The fields of real and complex numbers are denoted by $\mathbb{R}$ and $\mathbb{C}$, respectively. For a $3$rd-order tensor $\bm{\mathcal{A}}=[\bm{\mathcal{A}}_{ijk}]\in \mathbb{C}^{n_1\times n_2\times n_3}$, $\bm{\mathcal{A}}(i,:,:)$, $\bm{\mathcal{A}}(:,i,:)$ and $\bm{\mathcal{A}}(:,:,i)$ refer to the $i$th horizontal, lateral and frontal slice, respectively. In particular, ${\bm{\mathcal{{A}}}}^{(i)}$ stands for the frontal slice $\bm{\mathcal{A}}(:,:,i)$, and the complex conjugate of $\bm{\mathcal{A}}$ is denoted by $\rm conj(\bm{\mathcal{A}})$. The Frobenius inner product of $\bm A$ and $\bm B$ with the same dimensions is $\left\langle \bm A, \bm B\right\rangle = {\rm trace}(\bm A^T\bm B)$, and the inner product of $\bm{\mathcal{A}}$ and $\bm{\mathcal{B}}$ in $\mathbb{C}^{n_1\times n_2\times n_3}$ is $\left\langle \bm{\mathcal{A}}, \bm{\mathcal{B}}\right\rangle = \sum_{i=1}^{n_3} \left\langle {\bm{\mathcal{A}}} ^{(i)}, {\bm{\mathcal{B}}}^{(i)}\right\rangle$. Thus, the Frobenius and $\ell_\infty$ norms of $\bm{\mathcal{A}}$ are $\|\bm{\mathcal{A}}\|_F = \sqrt{\sum _{i=1}^{n_1}\sum_{j=1}^{n_2}\sum_{k=1}^{n_3} \bm{\mathcal{A}}_{ijk}^2}$ and $\|\bm{\mathcal{A}}\|_\infty = \max_{ijk} |\bm{\mathcal{A}}_{ijk}|$, respectively, and when $n_3=1$, $\|\bm{\mathcal{A}}\|_F$ becomes the Frobenius norm of a matrix.
	${\bm{\mathcal{\bar{A}}}}={\rm fft}(\bm{\mathcal{A}},[~],3)$ stands for the discrete Fourier transform (DFT) on each tube $\bm{\mathcal{A}}(i,j,:)$ via the MATLAB command $\rm fft$, while ${\bm{\mathcal{A}}}={\rm ifft}({\bm{\mathcal{\bar{A}}}},[~],3)$ denotes the inverse DFT on $\bm{\mathcal{A}}$.
	Furthermore, $\rm unfold(\bm{\mathcal{A}}) = [{\bm{\mathcal{A}}}^{(1)}; {\bm{\mathcal{A}}}^{(2)};\cdots;{\bm{\mathcal{A}}}^{(n_3)}]$ converts $\bm{\mathcal{A}}$ into a matrix with dimensions $n_1n_3\times n_2$, while $\rm fold(\cdot)$ is its inverse operation. The $\rm bdiag(\cdot)$ and $\rm bcirc(\cdot)$ are defined as:
	\begin{equation*}
		\bm{\bar{A}}={\rm bdiag}({\bm{\mathcal{\bar{A}}}}) = 
		\begin{bmatrix}
		{\bm{\mathcal{\bar{A}}}}^{(1)}&& &\\
			&{\bm{\mathcal{\bar{A}}}}^{(2)}&&\\
			&&\ddots &\\
			&&&{\bm{\mathcal{\bar{A}}}}^{(n_3)}
		\end{bmatrix}
	\end{equation*}
   and
	\begin{equation*}
		{\rm bcirc}(\bm{\mathcal{A}}) = 
		\begin{bmatrix}
			{\bm{\mathcal{A}}}^{(1)}&{\bm{\mathcal{A}}}^{(n_3)}&\cdots &{\bm{\mathcal{A}}}^{(2)}\\
			{\bm{\mathcal{A}}}^{(2)}&{\bm{\mathcal{A}}}^{(1)}&\cdots &{\bm{\mathcal{A}}}^{(3)}\\
			\vdots&\vdots&\ddots &\vdots\\
			{\bm{\mathcal{A}}}^{(n_3)}&{\bm{\mathcal{A}}}^{(n_3-1)}&\cdots &{\bm{\mathcal{A}}}^{(1)}
		\end{bmatrix}
	\end{equation*}
	Finally, $ |a|$ represents the absolute value of the scalar $a$.	
	The first and second derivatives of a differentiable function $f(x)$ are denoted by $f'(x)$ and $f''(x)$, respectively.
	
	\subsection{Preliminary Definitions}
	
	\begin{myDef}[t-product~\cite{KilmerEM2011}]
		Let $\bm{\mathcal{A}}\in \mathbb{R}^{n_1\times n_2\times n_3}$ and $\bm{\mathcal{B}}\in \mathbb{R}^{n_2\times l\times n_3}$. The t-product $\bm{\mathcal{A}}*\bm{\mathcal{B}}$ is the tensor $\bm{\mathcal{Z}}\in \mathbb{R}^{n_1\times l\times n_3}$ calculated by
		\begin{equation}
			\bm{\mathcal{Z}} = \bm{\mathcal{A}}*\bm{\mathcal{B}} = \rm fold(bcirc(\bm{\mathcal{A}})unfold(\bm{\mathcal{B}}))
		\end{equation}
	\end{myDef}
	\begin{myDef}[Identity tensor and f-diagonal tensor~\cite{KilmerEM2011}]
		The identity tensor $\bm{\mathcal{I}}\in \mathbb{R}^{n\times n\times n_3}$ is the tensor with its first frontal slice being an identity matrix and other frontal slices being all zeros. In particular, when each frontal slice is a diagonal matrix, the tensor is called f-diagonal.
	\end{myDef}
	\begin{myDef}[Conjugate transpose and orthogonal tensor~\cite{KilmerEM2011}]
		The conjugate transpose of a tensor $\bm{\mathcal{A}}\in \mathbb{C}^{n_1\times n_2\times n_3}$, referred to as $\bm{\mathcal{A}}^T\in \mathbb{C}^{n_2\times n_1\times n_3}$, is given by transposing each frontal slice and then reversing the order of transposed frontal slices $2$ through $n_3$. A tensor $\bm{\mathcal{Q}}\in \mathbb{R}^{n\times n\times n_3}$ is orthogonal if $\bm{\mathcal{Q}}^T*\bm{\mathcal{Q}} = \bm{\mathcal{Q}}*\bm{\mathcal{Q}}^T=\bm{\mathcal{I}}$.
	\end{myDef}
	\begin{myDef}[t-SVD, Theorem 2.2 in~\cite{LCYPAMItensor2019}]
		The tensor singular value decomposition (t-SVD) of a tensor $\bm{\mathcal{A}}\in \mathbb{R}^{n_1\times n_2\times n_3}$ is defined as:
		\begin{equation}
			\bm{\mathcal{A}}= \bm{\mathcal{U}}*\bm{\mathcal{S}}*\bm{\mathcal{V}}^T
		\end{equation}
		where $\bm{\mathcal{U}}\in \mathbb{R}^{n_1\times n_1\times n_3}$ and $\bm{\mathcal{V}}\in \mathbb{R}^{n_2\times n_2\times n_3}$ are orthogonal tensors and $\bm{\mathcal{S}}\in \mathbb{R}^{n_1\times n_2\times n_3}$ is an f-diagonal tensor.
	\end{myDef}
	It is known that the t-SVD of a tensor can be efficiently calculated in the Fourier domain as shown in Algorithm~\ref{Algo:t-SVD}~\cite{LCYPAMItensor2019}.
	\begin{algorithm}[htb]
		\caption{t-SVD}
		\label{Algo:t-SVD}
		\algsetup{indent=1.5em}
		\vspace{1ex}
		\begin{algorithmic}
			\REQUIRE  $\bm{\mathcal{A}} \in \mathbb{R}^{n_1\times n_2 \times n_3}$
			\begin{itemize}
				\item[1.]Compute $\bm{\mathcal{\bar{A}}}={\rm fft}(\bm{\mathcal{A}},[~],3)$
				\item[2.]Compute each frontal slice of $\bm{\mathcal{\bar{U}}}$, $\bm{\mathcal{\bar{S}}}$, and $\bm{\mathcal{\bar{V}}}$ from $\bm{\mathcal{\bar{A}}}$:		
				\FOR {$k=1,2,\cdots,\left \lceil \frac{n_3+1}{2}\right \rceil$}
				\STATE  $[{\bm{\mathcal{\bar{U}}}}^{(k)},{\bm{\mathcal{\bar{S}}}}^{(k)},{\bm{\mathcal{\bar{V}}}}^{(k)}] = {\rm SVD}({\bm{\mathcal{\bar{A}}}}^{(k)})$;
				\ENDFOR
				\FOR {$k=\left \lceil \frac{n_3+1}{2}\right \rceil+1,\cdots,n_3$}
				\STATE ${\bm{\mathcal{\bar{U}}}}^{(k)} = {\rm conj}({\bm{\mathcal{\bar{U}}}}^{(n_3-k+2)})$;
				\STATE ${\bm{\mathcal{\bar{S}}}}^{(k)} = {\bm{\mathcal{\bar{S}}}}^{(n_3-k+2)}$;
				\STATE ${\bm{\mathcal{\bar{V}}}}^{(k)} = {\rm conj}({\bm{\mathcal{\bar{V}}}}^{(n_3-k+2)})$;
				\ENDFOR
				\item[3.]Compute ${\bm{\mathcal{U}}}={\rm ifft}(\bm{\mathcal{\bar{U}}},[~],3)$, ${\bm{\mathcal{S}}}={\rm ifft}(\bm{\mathcal{\bar{S}}},[~],3)$, and ${\bm{\mathcal{V}}}={\rm ifft}(\bm{\mathcal{\bar{V}}},[~],3)$.
			\end{itemize}			
			\ENSURE ${\bm{\mathcal{U}}}$, ${\bm{\mathcal{S}}}$ and ${\bm{\mathcal{V}}}$.
		\end{algorithmic}
	\end{algorithm}
	\begin{myDef}[Tensor nuclear norm~\cite{LCYPAMItensor2019}]
		The tensor nuclear norm (TNN) of a tensor $\bm{\mathcal{A}}\in \mathbb{R}^{n_1\times n_2\times n_3}$ is given by:
		\begin{equation}\label{TNN}
			\|\bm{\mathcal{A}}\|_* = \frac{1}{n_3}\sum_{k=1}^{n_3}\|{\bm{\mathcal{\bar{A}}}}^{(k)}\|_* = \frac{1}{n_3}\sum_{k=1}^{n_3}\sum_{i=1}^{ r_k}\sigma_i({\bm{\mathcal{\bar{A}}}}^{(k)}) 
		\end{equation}
		where $ r_k \leq \min\{n_1,n_2\}$ is the rank of ${\bm{\mathcal{\bar{A}}}}^{(k)}$ and $\sigma_i({\bm{\mathcal{\bar{A}}}}^{(k)})$ is the $i$th singular value of ${\bm{\mathcal{\bar{A}}}}^{(k)}$.
	\end{myDef}
	\begin{lemma}[Relevant properties\cite{LCYPAMItensor2019}]\label{Im_properties_DFT} 
		There are two important properties for $\bm{\mathcal{A}}$ in the Fourier domain:
		\begin{equation*}
			\begin{split}
				\left<\bm{\mathcal{A}}, \bm{\mathcal{B}}\right> = \frac{1}{n_3}\left<\bm{\bar{A}}, \bm{\bar{B}}\right>,~~~
				\left\|\bm{\mathcal{A}}\right\|_F^2 = \frac{1}{n_3}\left\|\bm{\bar{A}}\right\|_F^2
			\end{split}
		\end{equation*}
	\end{lemma}
	\begin{lemma}[GSVT\cite{LuC2015}]\label{proximal_Matrix_norm0} 
		Let $\bm X=\bm U~{\rm Diag}(\bm s)~\bm V^T$ be the SVD of a rank-$r$ matrix $\bm X \in \mathbb{R}^{m\times n}$, where $\bm s = [s_1, s_2,\cdots,s_r]^T$ is the vector of singular values, and define:
		\begin{equation}\label{proximal_Matrix_norm3}
			{\bm {\hat{Y}}} = {\rm arg}\mathop {\min}\limits_{\bm Y}\lambda\|\bm Y\|_{\varphi} + \frac{1}{2}\left\|\bm X -\bm Y\right\|_F^2
		\end{equation}
	where $\|\bm Y\|_{\varphi}= \sum_{i=1}^{r}\varphi(\sigma_i(\bm Y))$.
	
		If the proximity operator $P_{\varphi}$ is monotonically non-decreasing, then the solution to (\ref{proximal_Matrix_norm3}) is:
		\begin{equation*}
			{\bm {\hat{Y}}}= \bm U {\rm Diag}(\bm s^\star)\bm V^T
		\end{equation*}
		where $\bm s^\star$ satisfies $s_1^\star\geq \cdots\geq s_i^\star\geq \cdots \geq s_r^\star $, which is determined for  $i = 1,2,\cdots,r$, as:  
		\begin{equation*}
			s_i^\star:= P_{\varphi}(s_i)={\rm arg}\mathop {\min}\limits_{s>0} \lambda\varphi(s) + \frac{1}{2}\left(s -s_i\right)^2
		\end{equation*}
	\end{lemma}
	
	\subsection{Related Works}
	\subsubsection{Low-Rank Tensor Completion}
	Given an observed tensor $\bm{\mathcal{X}}_\Omega$ with missing entries where $\Omega$ is the index set of the non-zero entries, namely, $\left(\bm{\mathcal{X}}_{\Omega}\right)_{ijk}= \bm{\mathcal{X}}_{ijk}$ if the index $\{i,j,k\}\in \Omega$, otherwise $\bm{\mathcal{X}}_{ijk} = 0$, the task of LRTC is to complete $\bm{\mathcal{X}}_\Omega$ using the low-rank property, which can be formulated as:
	\begin{equation}\label{TCrank_minimization}
		\mathop {\min}\limits_{\bm{\mathcal{M}}}~ \text{rank}(\bm{\mathcal{M}}), ~\text{s.t.} ~ \bm{\mathcal{M}}_{\Omega} = \bm{\mathcal{X}}_{\Omega}
	\end{equation}
	However, (\ref{TCrank_minimization}) is an NP-hard problem, and to solve it, many attempts have been exploited. Zhang $et$ $al.$~\cite{ZhangZETC2017} replace the rank minimization problem with a TNN minimization problem, resulting in
	\begin{equation}
		\mathop {\min}\limits_{\bm{\mathcal{M}}}~ \|\bm{\mathcal{M}}\|_{\text{TNN}}, ~\text{s.t.} ~ \bm{\mathcal{M}}_{\Omega} = \bm{\mathcal{X}}_{\Omega}
	\end{equation}
	Lu $et$ $al.$~\cite{LCYPAMItensor2019} solve the LRTC problem via a new TNN, leading to
	\begin{equation}
		\mathop {\min}\limits_{\bm{\mathcal{M}}}~ \|\bm{\mathcal{M}}\|_*, ~\text{s.t.} ~ \bm{\mathcal{M}}_{\Omega} = \bm{\mathcal{X}}_{\Omega}
	\end{equation}
	In fact, both apply a soft-thresholding operator to the singular value tensor in the Fourier domain. It is known that the soft-thresholding operator results in a biased solution. To alleviate the bias, nonconvex regularizers have been suggested~\cite{WangHIRTNN2022,YangMlogdet2022}:
	\begin{equation}
		\mathop {\min}\limits_{\bm{\mathcal{M}}}~ \|\bm{\mathcal{M}}\|_{\varphi}, ~\text{s.t.} ~ \bm{\mathcal{M}}_{\Omega} = \bm{\mathcal{X}}_{\Omega}
	\end{equation}
	where $\|\bm{\mathcal{M}}\|_{\varphi} = \frac{1}{n_3}\sum_{k=1}^{n_3}\sum_{i=1}^{ r_i}{\varphi}\left(\sigma_{ki}\right)$ with $\varphi(\cdot)$ being a nonconvex regularization function and $\sigma_{ki}$ is the $i$th singular value for the $k$th frontal slice of $\bm{\mathcal{\bar{M}}}$.
	
	\subsubsection{Half-Quadratic Optimization}
	Half-quadratic optimization (HO) was first proposed by Geman and Yang~\cite{GemanD1995}, which is used to optimize nonlinear functions via solving a sum of convex subproblems.
	Consider a function $\phi(x)$ such that $g(x)= x^2/2-\phi(x)$ is a closed proper convex function. Defining $g^*$ as the conjugate of $g$ and $\varphi(y)=g^*(y)-y^2/2$, we have:
	\begin{equation}
		\begin{split}
			g^*(y) =  \mathop {\max}\limits_{x} ~ x y - g(x)
		\end{split}
	\end{equation}
	\begin{equation}
		\varphi(y)=\mathop {\max}\limits_{x} ~ \phi(x)-\frac{1}{2}(x-y)^2
	\end{equation}
	Since $g(x)$ is convex, reciprocally, we obtain:
	\begin{equation}
		\begin{split}
			g(x) =  \mathop {\max}\limits_{y} ~ x y - g^*(y)
		\end{split}
	\end{equation}
	\begin{equation}\label{phi_minimization}
		\phi(x)=\mathop {\min}\limits_{y} ~ \frac{1}{2}(x-y)^2+ \varphi(y)
	\end{equation}
	According to the duality theory~\cite{RockafellarRT2004}, if $\phi(x)$ is differentiable, the solution to (\ref{phi_minimization}) is:
	\begin{equation}\label{soltion_HO}
		y=g'(x)=x-\phi'(x)
	\end{equation}
	where $g'(x)$ and $\phi'(x)$ are first order derivatives with respect to (w.r.t.) $x$ of $g(x)$ and $\phi(x)$, respectively.
	The above development also corresponds to the additive form of HO, and we call it HO in this work for convenience. For more details about HO, the interested reader is referred to~\cite{NikolovaM2005,IdierJHQ2001}. HO has been widely used in nonconvex function optimization in signal and image processing as well as machine learning. For example, He $et$ $al.$~\cite{HeRan2014,HeR2014} convert the Welsch function $\phi_{\rm welsch}=\frac{\sigma^2}{2}(1-{\rm exp}(-x^2/\sigma^2))$ into the following equivalent expression according to HO theory:
	\begin{equation}\label{Welsch_reg}
		\phi_{\rm welsch}=\mathop {\min}\limits_{y} ~ \frac{1}{2}(x-y)^2+ \varphi_{welsch}(y)
	\end{equation} 
	According to (\ref{soltion_HO}), the solution is:
	\begin{equation}\label{HO_welsch}
		y=  x-xe^{-x^2/\sigma^2}
	\end{equation}
 	\subsubsection{Proximity Operator/Thresholding Function}
 	The $Moreau~envelope$ for a proper and lower semicontinuous (lsc) regularizer $\varphi(\cdot)$ is defined as~\cite{CombettesP2005,BauschkeHH2011}:
	\begin{equation}\label{Def_Pro}
		\begin{split}
			\min\limits_{y}~\frac{1}{2}(x-y)^2 + \lambda\varphi(y)
		\end{split}    	   	
	\end{equation}
	whose solution is given by the proximity operator/thresholding function:
	\begin{equation}\label{R-LSp}
		\begin{split}
			P_\varphi(x) := {\rm \arg}\min\limits_{y}~\frac{1}{2}(x-y)^2 + \lambda\varphi(y)
		\end{split}    	   	
	\end{equation}
	If the regularizer $\varphi(y)$ makes the solution $P_\varphi(x)$ sparse, it is called the sparsity-inducing regularizer.
	For instance, if $\varphi(\cdot)$ is the $\ell_1$-norm, its $Moreau~envelope$ is:
	\begin{equation}\label{Moreau-L1}
		\begin{split}
			\min\limits_{y}~\frac{1}{2}(x-y)^2 + \lambda|y|_1
		\end{split}    	   	
	\end{equation}
	whose solution is:
	\begin{equation}\label{Pro-L1}
		\begin{split}
			y={\rm max}\{0,|x|-\lambda\}{\rm sign}(x)
		\end{split}    	   	
	\end{equation}
	where $\lambda\geq 0$ is a thresholding parameter, and (\ref{Pro-L1})
	is called the proximity operator of $|\cdot|_1$, also known as the soft-thresholding operator.
	
	\begin{figure}[htb]
		\centering
		\includegraphics[width=7cm]{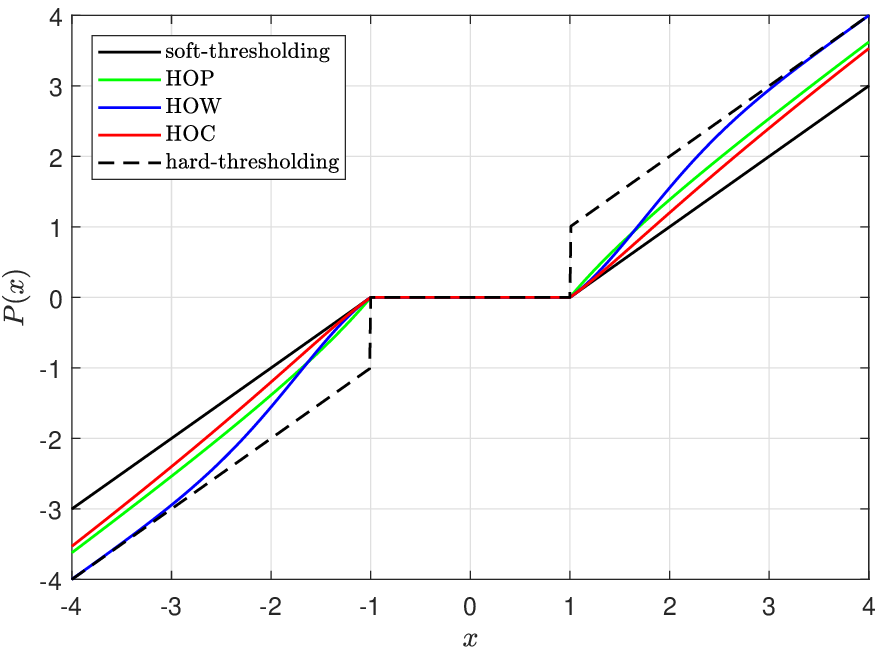}
		\vspace{-0.5em}
		\caption{Proximity operators with $\lambda=1$.}\label{Pro_Ope_comparison_TC}
	\end{figure}

	\section{Framework to Generate sparsity-inducing Regularizers}\label{Sec:framework}
	It is known that the $\ell_1$-norm is the sparsity-promoting regularizer and is able to make the solution sparse because $y=0$ for $|x|\leq \lambda$ for (\ref{Moreau-L1}). While for the regularizer generated by the Welsch function, it is not a sparsity-inducing regularizer and cannot make the solution to (\ref{Welsch_reg}) sparse since $y=0$ if and only if $x=0$ (see (\ref{HO_welsch})).
	According to (\ref{soltion_HO}), if we want to make the solution sparse, namely, $y=0$ for $|x|\leq \lambda$ like the $\ell_1$-norm, $\phi'(x)=x$ for $|x|\leq \lambda$, that is, $\phi(x)=x^2/2$ for $|x|\leq \lambda$. As an illustration, we consider the Huber function~\cite{HeRan2014}:
	\begin{equation}\label{Def_Huber1}
		\phi_{\rm huber}(x)=\begin{cases}
			x^2/2, &|x|\leq \lambda\\
			\lambda|x|-\frac{\lambda^2}{2}, &|x|\textgreater \lambda
		\end{cases}
	\end{equation}
	By HO, (\ref{Def_Huber1}) can be rewritten as:
	\begin{equation}\label{Moreau-huber}
		\begin{split}
			\phi_{\rm huber}(x) = \min\limits_{y}~\frac{1}{2}(x-y)^2 + \varphi_{\rm huber}(y)
		\end{split}    	   	
	\end{equation}
	and the solution to $y$ is given by:
	\begin{equation}\label{prox_huber}
		y=x-\phi_{\rm huber}'(x)=\begin{cases}
			0, &|x|\leq \lambda\\
			x-\lambda \cdot{\rm sign}(x), &|x|\textgreater \lambda
		\end{cases}
	\end{equation}
	Clearly, the regularizer $\varphi_{\rm huber}(y)$ is able to make the solution sparse. In particular, the regularizer generated by the Huber function is the $\ell_1$-norm~\cite{NikolovaM2005,HeRan2014}, while (\ref{Pro-L1}) and (\ref{prox_huber}) are equivalent.
	 
	As advocated in~\cite{FanJing2001}, the regularizers should make their proximity operators to achieve three properties: unbiasedness for large variable, sparsity, and continuity. As shown in Fig.~\ref{Pro_Ope_comparison_TC}, unlike the hard-thresholding operator that is discontinuous at $x=\lambda$, the soft-thresholding operator is continuous. Besides, the proximity operator of the $\ell_1$-norm produces a sparse solution at the expense of biasedness. In the estimation of sparse signals, the bias is the estimation error of sparse components~\cite{ZhangCHMCP2010,MeinshausenN2009}, which is determined as the gap $\Delta g$ between the identity function ($y=x$) and the proximity operator, namely, $\Delta g(x)=|x-P_\varphi(x)|$ for $|x|\geq \lambda$. From (\ref{Pro-L1}), the bias generated by the $\ell_1$-norm is $\lambda$ when $x>\lambda$. Note that we only discuss the case for $x>\lambda$ since the proximity operator is an odd function.
	To obtain an unbiased solution, the gap should decrease as the variable increases, that is, $\Delta g(x_2)\leq \Delta g(x_1)$ for $x_2\geq x_1\geq \lambda$:
	\begin{equation}\label{bias_gap}
		x_2-P_\varphi(x_2) \leq x_1 -P_\varphi(x_1) 
	\end{equation}
	where $P_\varphi(x) = x - \phi'(x)$ via (\ref{soltion_HO}).
	This results in
	\begin{equation}\label{phi_concave}
		\phi'(x_2)\leq \phi'(x_1)
	\end{equation}
	implying that $\phi''(x)\leq 0$ and $\phi(x)$ is concave for $x>\lambda$. Accordingly, we state the following proposition.
	\begin{mypro}
		Consider a differentiable loss function $\phi_{h,\lambda}(x)$ such that $g(x)=x^2/2-\phi_{h,\lambda}(x)$ is a closed, proper, convex function. If $\phi_{h,\lambda}(x)=x^2/2$ for $|x|\leq \lambda$, and $\phi_{h,\lambda}(x)= a\cdot h(|x|)+b$ is concave for $|x|>\lambda$ where $a$ and $b$ are constants to make $\phi_{h,\lambda}(x)$ continuously differentiable, then it can be used to generate a sparsity-inducing regularizer $\varphi_{h,\lambda}(\cdot)$ via HO, that is,
		\begin{equation}\label{covex_newphi}
			\phi_{h,\lambda}(x)=\mathop {\min}\limits_{y} ~ \frac{1}{2}(x-y)^2+ \lambda\varphi_{h,\lambda}(y)
		\end{equation}
	where $\varphi_{h,\lambda}(y)=\mathop {\max}\limits_{x} ~ \phi_{h,\lambda}(x)/\lambda-\frac{1}{2\lambda}(x-y)^2$ is the sparsity-promoting regularizer.
	The solution to $y$ in (\ref{covex_newphi}) is:
	\begin{equation}
		P_{\varphi_{h,\lambda}}(x)= x- \phi'_{h,\lambda}(x)= {\rm max}\left\{0, |x|-a\cdot h'(|x|) \right\}\cdot {\rm sign}(x)
	\end{equation}
	where $a$ is constant associated with $\lambda$.
	If $\phi_{h,\lambda}(x)$ is strictly concave for $x>\lambda$, the resultant proximity operator makes the solution have less bias than the proximity operator of the $\ell_1$-norm.
	\end{mypro}
	Proof: See Appendix A.
	
	It is worth noting that the exact expression of $\varphi_{h,\lambda}(\cdot)$ is generally unknown due to the property of HO~\cite{GemanD1995,NikolovaM2005,IdierJHQ2001}. 
	Nevertheless, in many cases, to avoid iterations, the closed-form expression of the proximity operator is desired while it is not necessary to know the exact expression of the regularizer. In addition, the properties of $\varphi_{h,\lambda}(\cdot)$ are analyzed in Proposition~\ref{Properties_phi}.
	\begin{mypro}\label{Properties_phi}
		The sparsity-inducing regularizer $\varphi_{h,\lambda}(\cdot)$ satisfies the following properties although its expression is generally unknown:
		\begin{itemize}
			\item[(i)] The $Moreau~envelope$ of $\varphi_{h,\lambda}(\cdot)$ is a convex problem, namely, (\ref{covex_newphi}) is a convex problem w.r.t. $y$.
			\item[(ii)] $\varphi_{h,\lambda}(y)$ with $y>0$ is concave if $\phi_{h,\lambda}(x)$ is concave for $x>\lambda$, and $\varphi'_{h,\lambda}(y)< 1$ for $y>0$.
			\item[(iii)] The thresholding function $P_{\varphi_{h,\lambda}}(x)$ is monotonically nondecreasing, and if $\phi_{h,\lambda}(x)$ is concave for $x>\lambda$, $P_{\varphi_{h,\lambda}}(x)$ increases with $x$ for $x>\lambda$.
		\end{itemize}
	\end{mypro}
	Proof: See Appendix B.
	
	Next, we generalize $h(x)$ via some commonly-used nonconvex functions.
	\subsection{Generalization via $\ell_p$-norm }
	When $h(x)=|x|^p$, to make $\phi(x)$ differentiable, we adopt the smooth hybrid ordinary-$\ell_p$ (HOP) function, where `ordinary' refers to the quadratic function:
	\begin{equation}\label{lp-ab}
		\phi_{p,\lambda}(x) = 
		\begin{cases}
			x^2/2, &|x|\leq \lambda\\
			\frac{1}{p}\lambda^{2-p} |x|^p+\frac{\lambda^2}{2}-\frac{1}{p}\lambda^2, &|x|\textgreater \lambda
		\end{cases}
	\end{equation}
	According to HO, we have 
	\begin{equation}\label{l-p-function}
		\begin{split}
			\phi_{p,\lambda}(x) = \mathop {\min}\limits_{y} ~\frac{(x-y)^2}{2} + \lambda\varphi_{p,\lambda}(y)
		\end{split}	
	\end{equation}
	where $\varphi_{p,\lambda}(y) $ is the sparse regularizer related to $\phi_{p,\lambda}(x)$,	and the solution to (\ref{l-p-function}) is:
	\begin{equation}\label{lp-pro-solution}
		P_{{\varphi}_{p,\lambda}}(x) =x-\phi_{p,\lambda}'(x)= {\rm max}\left\{0, |x|-{\lambda}^{2-p}|x|^{p-1} \right\}\cdot {\rm sign}(x)
	\end{equation}
	
	It is worth pointing out that when $p=1$, (\ref{lp-ab}) becomes the Huber function and (\ref{lp-pro-solution}) is equal to (\ref{Pro-L1}).
	
	\subsection{Generalization via Welsch function}
	
	Unlike $|x|^p$ with $0<p<1$ that is concave for $x>0$, the Welsch function is another type of function that is nonconvex but not concave when $x>0$. When $h(x)=\phi_{\rm welsch}(x)$, the hybrid ordinary-Welsch (HOW) function is given by:
	\begin{equation*}
		\phi_{\sigma,\lambda}(x) = 
		\begin{cases}
			x^2/2, &|x|\leq \lambda\\
			\frac{\sigma^2}{2}\left(1-e^{\frac{\lambda^2-x^2}{\sigma^2}}\right)+\frac{\lambda^2}{2}, &|x|\textgreater \lambda
		\end{cases}
	\end{equation*}
	where $\sigma$ is the kernel size. Using the HO results in:
	\begin{equation}\label{l-welsch-equl}
		\begin{split}
			\phi_{\sigma,\lambda}(x) = \mathop {\min}\limits_{y} ~\frac{(x-y)^2}{2} + \lambda\varphi_{\sigma,\lambda}(y)
		\end{split}	
	\end{equation}
	According to (\ref{soltion_HO}), the solution to (\ref{l-welsch-equl}) is given by:
	\begin{equation}\label{lw-pro-solution}
		P_{\varphi_{\sigma,\lambda}}(x) := {\rm max}\left\{0, |x|-|x|\cdot e^{(\lambda^2-x^2)/\sigma^2}\right\}\cdot {\rm sign}(x)
	\end{equation}
	It is worth mentioning that compared with (\ref{HO_welsch}), the proximity operator in (\ref{lw-pro-solution}) can yield a sparse solution because $P_{\varphi_{\sigma,\lambda}}(x)=0$ for $|x|\leq \lambda$. It is seen that HOP with $p<1$ is concave for $x>\lambda$ because $|x|^p$ with $p<1$ is concave for $x>0$, while HOW may not be concave for $x>\lambda$. Therefore, it is necessary to make sure that $\phi_{\sigma,\lambda}(x)$ is concave for $x>\lambda$. When $x>\lambda$, we have:
	\begin{equation*}
		\phi_{\sigma,\lambda}''(x)=\left(1-\frac{2x^2}{\sigma^2}\right)e^{\frac{\lambda^2-x^2}{\sigma^2}}\leq 0
	\end{equation*}
	resulting in $\sigma\leq \sqrt{2}x$. As $x>\lambda$, this leads to
	\begin{equation*}
		\sigma\leq \sqrt{2}\lambda
	\end{equation*}
	That is, when $\sigma\leq \sqrt{2}\lambda$, $\phi_{\sigma,\lambda}(x)$ is concave for $x>\lambda$ and the bias generated by the thresholding operator (\ref{lw-pro-solution}) is less than that by the soft-thresholding operator.
	\subsection{Generalization via Cauchy function}
	We now substitute $h(x)$ with the Cauchy function $\phi_{\rm Cauchy}(x)$, whose expression is:
	\begin{equation*}
		\phi_{\rm Cauchy}(x) =\ln \left(1+\left(\frac{x}{\gamma}\right)^2\right)
	\end{equation*}
	where $\gamma$ is the scale parameter. Then, we have the hybrid ordinary-Cauchy (HOC) function:
	\begin{equation*}
		\phi_{\gamma,\lambda}(x) = 
		\begin{cases}
			x^2/2, &|x|\leq \lambda\\
			\frac{\gamma^2+\lambda^2}{2}\ln \left(1+\left(\frac{x}{\gamma}\right)^2\right)+b, &|x|\textgreater \lambda
		\end{cases}
	\end{equation*}
	where $b=\frac{\lambda^2}{2}-\frac{\gamma^2+\lambda^2}{2}\ln \left(1+\left({\lambda}/{\gamma}\right)^2\right)$. Employing the HO results in:
	\begin{equation*}
		\begin{split}
			\phi_{\gamma,\lambda}(x) = \mathop {\min}\limits_{y} ~\frac{(x-y)^2}{2} + \lambda\varphi_{\gamma,\lambda}(y)
		\end{split}	
	\end{equation*}
	According to (\ref{soltion_HO}), the solution is:
	\begin{equation}\label{lc-pro-solution}
		p_{\varphi_{\gamma,\lambda}}(x) := {\rm max}\left\{0, |x|-\frac{\left(\gamma^2+\lambda^2\right)|x|}{\gamma^2+x^2}\right\}\cdot {\rm sign}(x)
	\end{equation}
	To make $\phi_{\gamma,\lambda}(x)$ concave for $x>\lambda$, we have
	\begin{equation*}
		\phi_{\gamma,\lambda}''(x)=\frac{(\gamma^2+\lambda^2)(\gamma^2-x^2)}{(\gamma^2+x^2)^2}\leq 0
	\end{equation*}
	leading to $\gamma\leq x$. As $x>\lambda$, we obtain:
	\begin{equation}
		\gamma\leq \lambda
	\end{equation}
	The curves of the proximity operator associated with the three devised loss functions, i.e., HOP, HOW and HOC, are also depicted in Fig.~\ref{Pro_Ope_comparison_TC}, where we set $\lambda=1$, $\sigma=\sqrt{2}\lambda$, $\gamma=\lambda$ and $p=0.3$ for HOP. It is seen that compared with the soft-thresholding operator, our thresholding functions introduce less bias in the solution since their curves approach the identity function asymptotically.

	\section{Algorithm Development}\label{Sec:developed_algorithm}
	\subsection{Generalized Tensor Singular Value Thresholding}
	Applying TNN for low-rank tensor recovery will underestimate the nonzero singular values of frontal slices in the Fourier domain because TNN utilizes the $\ell_1$-norm as a penalty on the nonzero singular values. To address this issue, the $\ell_1$-norm is replaced by our sparsity-promoting regularizers.
	\begin{myDef}
		Similar to~(\ref{TNN}), the tensor ${\varphi_{h,\lambda}}$ norm of $\bm{\mathcal{A}}\in \mathbb{R}^{n_1\times n_2\times n_3}$, denoted by $\|\bm{\mathcal{A}}\|_{\varphi_{h,\lambda}}$, is defined as:
		\begin{equation}\label{tensor_phi_norm}
			\|\bm{\mathcal{A}}\|_{\varphi_{h,\lambda}} = \frac{1}{n_3}\sum_{k=1}^{n_3}\|\bm{\mathcal{\bar{A}}}^{(k)}\|_{\varphi_{h,\lambda}} = \frac{1}{n_3}\sum_{k=1}^{n_3}\sum_{i=1}^{ r_i}{\varphi_{h,\lambda}}\left(\sigma_i(\bm{\mathcal{\bar{A}}}^{(k)})\right) 
		\end{equation}
		where $ r_i \leq \min\{n_1,n_2\}$ is the rank of $\bm{\mathcal{\bar{A}}}^{(k)}$, $\sigma_i(\bm{\mathcal{\bar{A}}}^{(k)})$ is the $i$th singular value of $\bm{\mathcal{\bar{A}}}^{(k)}$, and $\varphi_{h,\lambda}(\cdot)$ is the sparsity-inducing regularizer, which is a nonconvex regularization function.
	\end{myDef}
	
	Besides, we provide the generalized tensor singular value thresholding (GTSVT) operator in the following theorem.
	\begin{theorem}[GTSVT]\label{GTSVT_TC}
		Let $\bm{\mathcal{X}}= \bm{\mathcal{U}}*\bm{\mathcal{S}}*\bm{\mathcal{V}}^T$ be the t-SVD of $\bm{\mathcal{X}}\in \mathbb{R}^{n_1\times n_2\times n_3}$ and define
		\begin{equation}\label{nonconvex_TNN_op}
			\mathcal{D}_{\varphi_{h,\lambda}} (\bm{\mathcal{X}}) = {\rm arg}\mathop {\min}\limits_{\bm{\mathcal{Y}}}\lambda\|\bm{\mathcal{Y}}\|_{\varphi_{h,\lambda}} + \frac{1}{2}\left\|\bm{\mathcal{X}}-\bm{\mathcal{Y}}\right\|_F^2
		\end{equation}
		If the proximity operator $P_{\varphi_{h,\lambda}}$ is monotonically non-decreasing, then the solution to (\ref{nonconvex_TNN_op}) is:
		\begin{equation}
		\bm{\mathcal{Y}}=\bm{\mathcal{U}}*\bm{\mathcal{S}}_{\varphi_{h,\lambda}}*\bm{\mathcal{V}}^T
		\end{equation}
		where $\bm{\mathcal{S}}_{\varphi_{h,\lambda}}= {\rm ifft}\left(P_{\varphi_{h,\lambda}}(\bm{\mathcal{\bar{S}}}),[~],3\right)$.
	\end{theorem}
	Proof: See Appendix C.
	
	Based on Algorithm~\ref{Algo:t-SVD}, the GTSVT procedure is provided in Algorithm~\ref{Algo:GTSVT}. The difference between the GTSVT and t-SVT~\cite{LCYPAMItensor2019} is the proximal operator on $\bm{\mathcal{\bar{S}}}$, and the latter adopts the soft-thresholding to the singular value of $\bm{\mathcal{\bar{S}}}$.
	\begin{algorithm}[htb]
		\caption{GTSVT}
		\label{Algo:GTSVT}
		\algsetup{indent=1.5em}
		\vspace{1ex}
		\begin{algorithmic}
			\REQUIRE  $\bm{\mathcal{X}} \in \mathbb{R}^{n_1\times n_2 \times n_3}$ and $\lambda>0$
			\begin{itemize}
				\item[1.]Compute $\bm{\mathcal{\bar{X}}}={\rm fft}(\bm{\mathcal{X}},[~],3)$
				\item[2.]Perform GSVT on all frontal slices of $\bm{\mathcal{\bar{X}}}$:
				\FOR {$k=1,2,\cdots,\left \lceil \frac{n_3+1}{2}\right \rceil$}
				\STATE  $[\bm{\mathcal{\bar{U}}}^{(k)},\bm{\mathcal{\bar{S}}}^{(k)},\bm{\mathcal{\bar{V}}}^{(k)}] = {\rm SVD}(\bm{\mathcal{\bar{X}}}^{(k)})$;
				\STATE $\bm{\mathcal{\bar{Y}}}^{(k)} = \bm{\mathcal{\bar{U}}}^{(k)} P_{\varphi_{h,\lambda}}(\bm{\mathcal{\bar{S}}}^{(k)})(\bm{\mathcal{\bar{V}}})^T$;
				\ENDFOR
				\FOR {$k=\left \lceil \frac{n_3+1}{2}\right \rceil+1,\cdots,n_3$}
				\STATE $\bm{\mathcal{\bar{Y}}}^{(k)} = {\rm conj}(\bm{\mathcal{\bar{Y}}}^{(n_3-k+2)})$;
				\ENDFOR
				\item[3.]Compute $\mathcal{D}_{\varphi_{h,\lambda}} (\bm{\mathcal{X}})={\rm ifft}(\bm{\mathcal{\bar{Y}}},[~],3)$
			\end{itemize}			
			\ENSURE $\mathcal{D}_{\varphi_{h,\lambda}} (\bm{\mathcal{X}})$.
		\end{algorithmic}
	\end{algorithm}

	\subsection{Algorithm Development}
	The LRTC problem can be written as:
	\begin{equation}
		\begin{split}
			\min\limits_{\bm{\mathcal{M}}}~\|\bm{\mathcal{M}}\|_{\varphi_{h,1/\rho}}, ~ \text{s.t.} ~ \bm{\mathcal{M}}_\Omega = \bm{\mathcal{X}}_\Omega
		\end{split}  
	\end{equation}
	which is equal to:
	\begin{equation}\label{TC_model}
		\begin{split}
			\min\limits_{\bm{\mathcal{M}}}~\|\bm{\mathcal{M}}\|_{\varphi_{h,1/\rho}}, ~ \text{s.t.} ~ \bm{\mathcal{M}} + \bm{\mathcal{E}} = \bm{\mathcal{X}}, ~~\bm{\mathcal{E}}_\Omega=\pmb 0
		\end{split}
	\end{equation}
	where $\bm{\mathcal{E}}_{ \Omega^c} \neq \pmb 0$ if $\bm{\mathcal{M}}_{ \Omega^c} \neq \pmb 0$. Problem (\ref{TC_model}) can be efficiently solved by ADMM, and its augmented Lagrangian function is:
	\begin{equation}
		\begin{split}
			\mathcal{L}'_\rho(\bm{\mathcal{M}}, \bm{\mathcal{E}},\bm{\mathcal{P}}) = &\|\bm{\mathcal{M}}\|_{\varphi_{h,1/\rho}} +\left<\bm{\mathcal{P}}, \bm{\mathcal{X}}-\bm{\mathcal{M}} -\bm{\mathcal{E}}\right> \\
			& +\frac{\rho}{2}\left\|\bm{\mathcal{X}}-\bm{\mathcal{M}} -\bm{\mathcal{E}}\right\|_F^2
		\end{split}
	\end{equation}
	which amounts to: 
	\begin{equation}\label{TC_Aug_L}
		\begin{split}
			\mathcal{L}_\rho(\bm{\mathcal{M}}, \bm{\mathcal{E}},\bm{\mathcal{P}}) = &\frac{1}{\rho}\|\bm{\mathcal{M}}\|_{\varphi_{h,1/\rho}} +\frac{1}{\rho}\left<\bm{\mathcal{P}}, \bm{\mathcal{X}}-\bm{\mathcal{M}} -\bm{\mathcal{E}}\right> \\
			& +\frac{1}{2}\left\|\bm{\mathcal{X}}-\bm{\mathcal{M}} -\bm{\mathcal{E}}\right\|_F^2
		\end{split}
	\end{equation}
	where $\bm{\mathcal{P}}$ is the Lagrange multiplier tensor and $\rho>0$ is the penalty parameter.
	Given the estimates at the $n$th iteration, namely, $\bm{\mathcal{E}}_n$, $\bm{\mathcal{M}}_n$ and $\bm{\mathcal{P}}_n$, their updating rules at the $(n+1)$th iteration are derived as follows:
	
	$Update~of $~$\bm{\mathcal{E}}$: Given $\bm{\mathcal{M}}_n$, $\bm{\mathcal{P}}_n$ and $\rho_n$, $\bm{\mathcal{E}}_{n+1}$ is calculated by:
	\begin{equation}\label{tensor_E_problem}
		\begin{split}
			{\rm arg}\mathop {\min}\limits_{\bm{\mathcal{E}}}~ \frac{1}{2}\left\|\bm{\mathcal{X}}-\bm{\mathcal{M}}_n + \frac{\bm{\mathcal{P}}_n}{\rho_n}-\bm{\mathcal{E}}\right\|_F^2~ \text{s.t.} ~\bm{\mathcal{E}}_\Omega=0
		\end{split}
	\end{equation}
	resulting in:
	\begin{equation}\label{TC_Esol}
		\begin{split}
			(\bm{\mathcal{E}}_{n+1})_{\Omega^c}= -(\bm{\mathcal{M}}_n)_{\Omega^c} + \frac{(\bm{\mathcal{P}}_{n})_{\Omega^c}}{\rho_n}
		\end{split}
	\end{equation}
	where $\Omega^c$ is the complementary set of $\Omega$.
	
	$Update~of $~$\bm{\mathcal{M}}$: Given $\bm{\mathcal{E}}_{n+1}$, $\bm{\mathcal{P}}_n$ and $\rho_n$, $\bm{\mathcal{M}}_{n+1}$ is determined as:
	\begin{equation}\label{sol_t_M}
		\begin{split}
			 {\rm arg}\mathop {\min}\limits_{\bm{\mathcal{M}}}~\frac{1}{\rho}\|\bm{\mathcal{M}}\|_{\varphi_{h,1/\rho}} +\frac{1}{2}\left\|\bm{\mathcal{X}}- \bm{\mathcal{E}}_{n+1} + \frac{\bm{\mathcal{P}}_n}{\rho_n} -\bm{\mathcal{M}} \right\|_F^2
		\end{split}
	\end{equation}
	Defining $\bm{\mathcal{R}}_{n} = \bm{\mathcal{X}}- \bm{\mathcal{E}}_{n+1} + \frac{\bm{\mathcal{P}}_n}{\rho_n}=\bm{\mathcal{U}}_{n}*\bm{\mathcal{S}}_{n}*(\bm{\mathcal{V}}_{n})^T$, and according to Theorem~\ref{GTSVT_TC}, the solution to (\ref{sol_t_M}) is:
	\begin{equation}\label{TC_Msol}
		\bm{\mathcal{M}}_{n+1} = \bm{\mathcal{U}}_{n}*(\bm{\mathcal{S}}_{\varphi_{h,1/{\rho_n}}})_{n}*(\bm{\mathcal{V}}_{n})^T
	\end{equation}
	$Update~of $~$\bm{\mathcal{P}}$: Given $\bm{\mathcal{M}}_{n+1}$, $\bm{\mathcal{E}}_{n+1}$ and $\rho_n$, we have
	\begin{equation}\label{TC_Psol}
		\bm{\mathcal{P}}_{n+1} = \bm{\mathcal{P}}_n+\rho_n\left(\bm{\mathcal{X}}-\bm{\mathcal{M}}_{n+1} -\bm{\mathcal{E}}_{n+1}\right)
	\end{equation}
	Besides, $\rho_{n+1}=\mu \rho_n$ with $\mu>1$. The entire iterative procedure is summarized in Algorithm~\ref{Algo:GTSVT-TC}. It is worth pointing out that ${\varphi_{h,1/\rho}}$ can be replaced by ${\varphi_{p,1/\rho}}$, ${\varphi_{\sigma,1/\rho}}$ and ${\varphi_{\gamma,1/\rho}}$, which are the respective regularizers generated by the HOP, HOW and HOC, and we denote their developed algorithms as GTNN-HOP, GTNN-HOW and GTNN-HOC, respectively. For example, if the regularizer ${\varphi_{p,1/\rho}}$ is adopted in Algorithm~\ref{Algo:GTSVT-TC}, we refer the resultant algorithm to as GTNN-HOP.
	\begin{algorithm}
		\caption{GTSVT based tensor completion}
		\label{Algo:GTSVT-TC}
		\algsetup{indent=1.5em}
		\vspace{1ex}
		\begin{algorithmic}
			\REQUIRE  Observed tensor $\bm{\mathcal{X}}_\Omega$, index set $\Omega$, $\xi>0$ and $I_m$
			\STATE \textbf{Initialize:} $\bm{\mathcal{M}}_0=\bm{\mathcal{X}}_\Omega$, $\bm{\mathcal{P}}_0=\pmb 0$, $\rho_0 =10^{-4}$, $\mu=1.2$, and $n=0$.
			\WHILE {not converged and $n\leq I_m$}
			
			\STATE Update $\bm{\mathcal{E}}_n$ via (\ref{TC_Esol})
			
			\STATE Update $\bm{\mathcal{M}}_n$ via (\ref{TC_Msol})	
			
			\STATE Update $\bm{\mathcal{P}}_n$ via (\ref{TC_Psol})
			
			\STATE Update $\rho_{n+1}=\mu\rho_n$	
			
			\STATE Check the convergence conditions
			\begin{equation*}
			\begin{split}
				&\left\|\bm{\mathcal{M}}_{n+1}-\bm{\mathcal{M}}_n\right\|_\infty \leq \xi,~\left\|\bm{\mathcal{E}}_{n+1}-\bm{\mathcal{E}}_n\right\|_\infty \leq \xi\\
				&\left\|\bm{\mathcal{X}}-\bm{\mathcal{M}}_{n+1} -\bm{\mathcal{E}}_{n+1}\right\|_\infty \leq \xi
			\end{split}
			\end{equation*}
						
			\STATE $k\leftarrow k+1$
			
			\ENDWHILE
			\ENSURE $\bm{\mathcal{M}}= \bm{\mathcal{M}}_n$.
		\end{algorithmic}
	\end{algorithm}
	
	\subsection{Convergence Analysis}
	The convergence of Algorithm~\ref{Algo:GTSVT-TC} is analyzed in the following theorem. 
	\begin{theorem}\label{theorem-convergence} 
		Let $\{\bm{\mathcal{M}}_n, \bm{\mathcal{E}}_n, \bm{\mathcal{P}}_n\}$ be the sequence generated by Algorithm~\ref{Algo:GTSVT-TC}. Given a bounded initialization, $\{\bm{\mathcal{M}}_n, \bm{\mathcal{E}}_n, \bm{\mathcal{P}}_n\}$ has the following properties:
		\begin{itemize}
			\item[(i)]The generated sequences $\{\bm{\mathcal{M}}_n, \bm{\mathcal{E}}_n, \bm{\mathcal{P}}_n\}$ are all bounded.
			\item[(ii)] The sequence $\{\bm{\mathcal{M}}_n, \bm{\mathcal{E}}_n\}$ satisfies:
			\begin{enumerate}
				\item
				$\lim_{n \to \infty}\left\|\bm{\mathcal{E}}_{n+1}-\bm{\mathcal{E}}_n\right\|_F^2=0$
				\item
				$\lim_{n \to \infty}\left\|\bm{\mathcal{M}}_{n+1}-\bm{\mathcal{M}}_n\right\|_F^2=0$
				\item
				$\lim_{n \to \infty}\left\|\bm{\mathcal{X}}-\bm{\mathcal{M}}_{n+1} -\bm{\mathcal{E}}_{n+1}\right\|_F^2=0$
			\end{enumerate}
			\item[(iii)]Any limit point $\{\bm{\mathcal{M}}_\star, \bm{\mathcal{E}}_\star, \bm{\mathcal{P}}_\star\}$ is a stationary point that satisfies the KKT conditions for (\ref{TC_model}). 
		\end{itemize}
	\end{theorem}
	Proof: See Appendix D.
	
	\subsection{Computational Complexity}
	
	The main computation cost of our algorithm lies in the update of $\bm{\mathcal{M}}_n\in \mathbb{R}^{n_1\times n_2\times n_3}$ per iteration. It includes computing the FFT along the third model with complexity $\mathcal{O}(n_1n_2n_3\log(n_3))$, the SVD of frontal slices with complexity $\mathcal{O}(n_1n_2n_3\min\{n_1,n_2\})$ and the inverse FFT of $\bm{\mathcal{\bar{M}}}_n$ with complexity $\mathcal{O}(n_1n_2n_3\log(n_3))$. Thus, the total complexity is $\mathcal{O}(2n_1n_2n_3\log(n_3)+n_1n_2n_3\min\{n_1,n_2\})$ per iteration.

	\begin{figure}[htb]
		\centering
		\begin{minipage}{0.23\linewidth}
			\footnotesize
			\vspace{1pt}
			\centerline{\includegraphics[width=4.2cm]{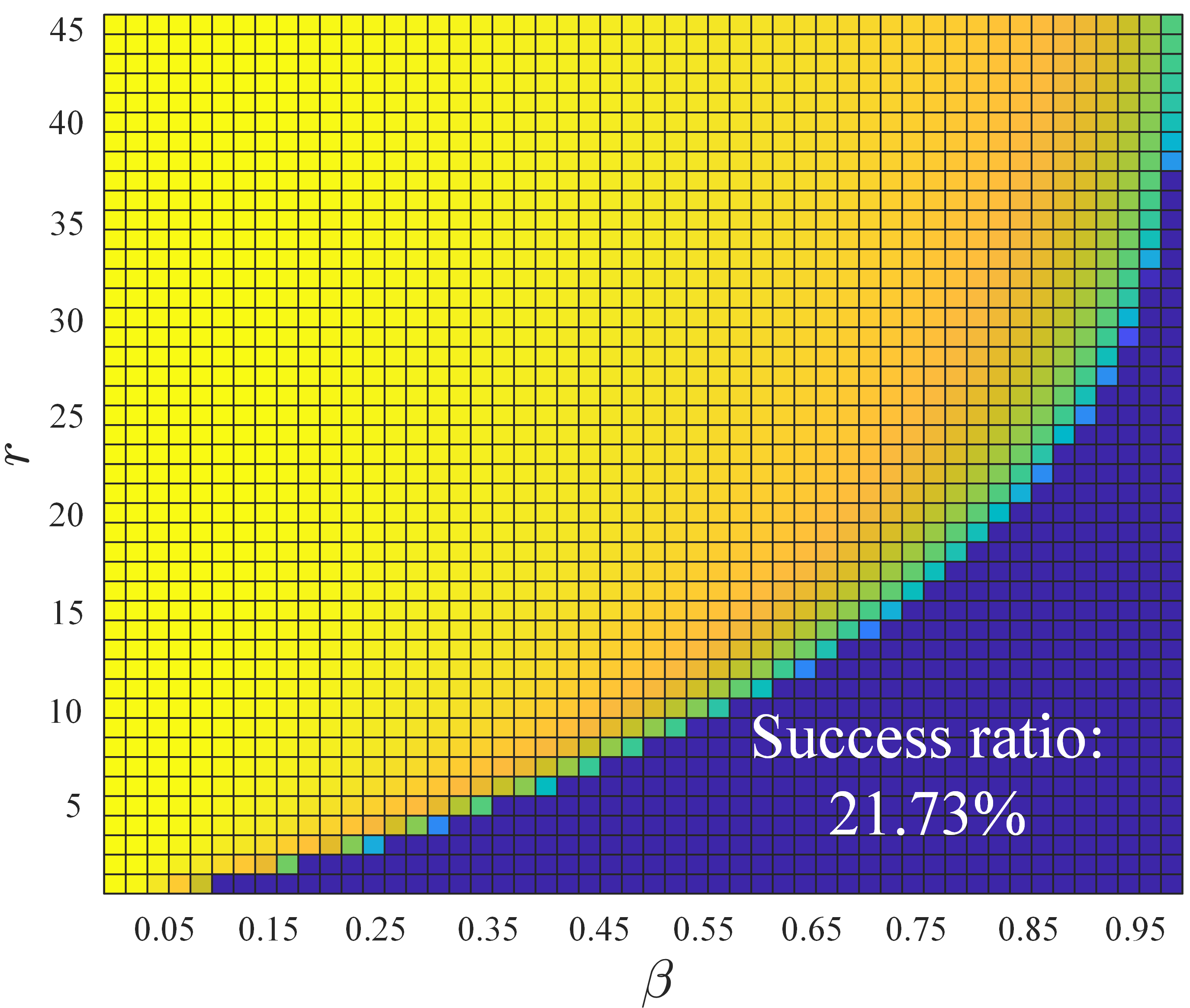}}\vskip 0pt
			\centerline{\scriptsize {(a). TNN}}\vskip -3pt
			\centerline{ }
		\end{minipage}\hspace{23mm}
		\begin{minipage}{0.23\linewidth}
			\footnotesize
			\vspace{1pt}
			\centerline{\includegraphics[width=4.7cm]{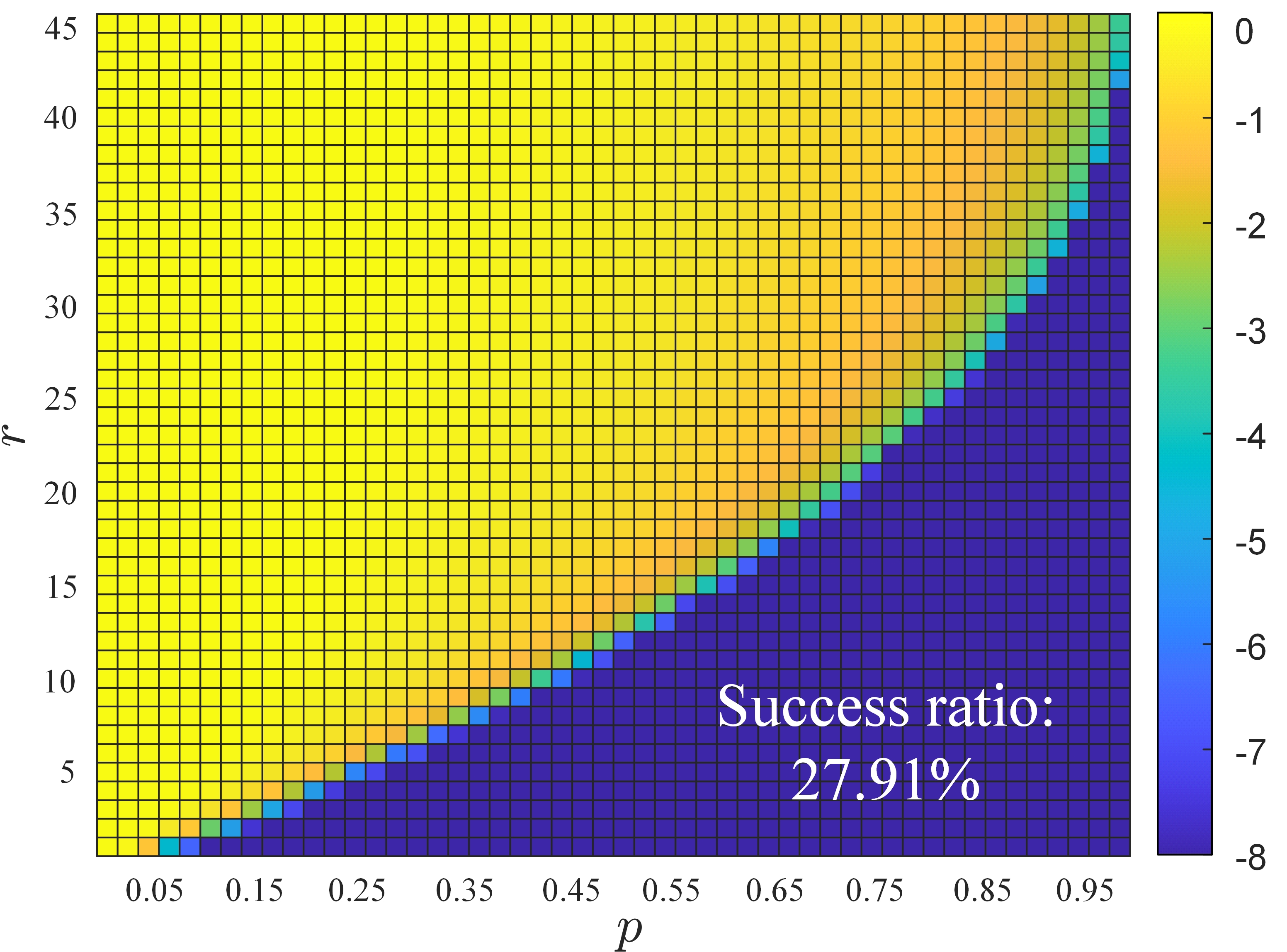}}\vskip 0pt
			\centerline{\scriptsize {(b). GTNN-HOP}}\vskip -3pt
			\centerline{ }
		\end{minipage}
		
		\begin{minipage}{0.23\linewidth}
			\footnotesize
			\vspace{1pt}
			\centerline{\includegraphics[width=4.2cm]{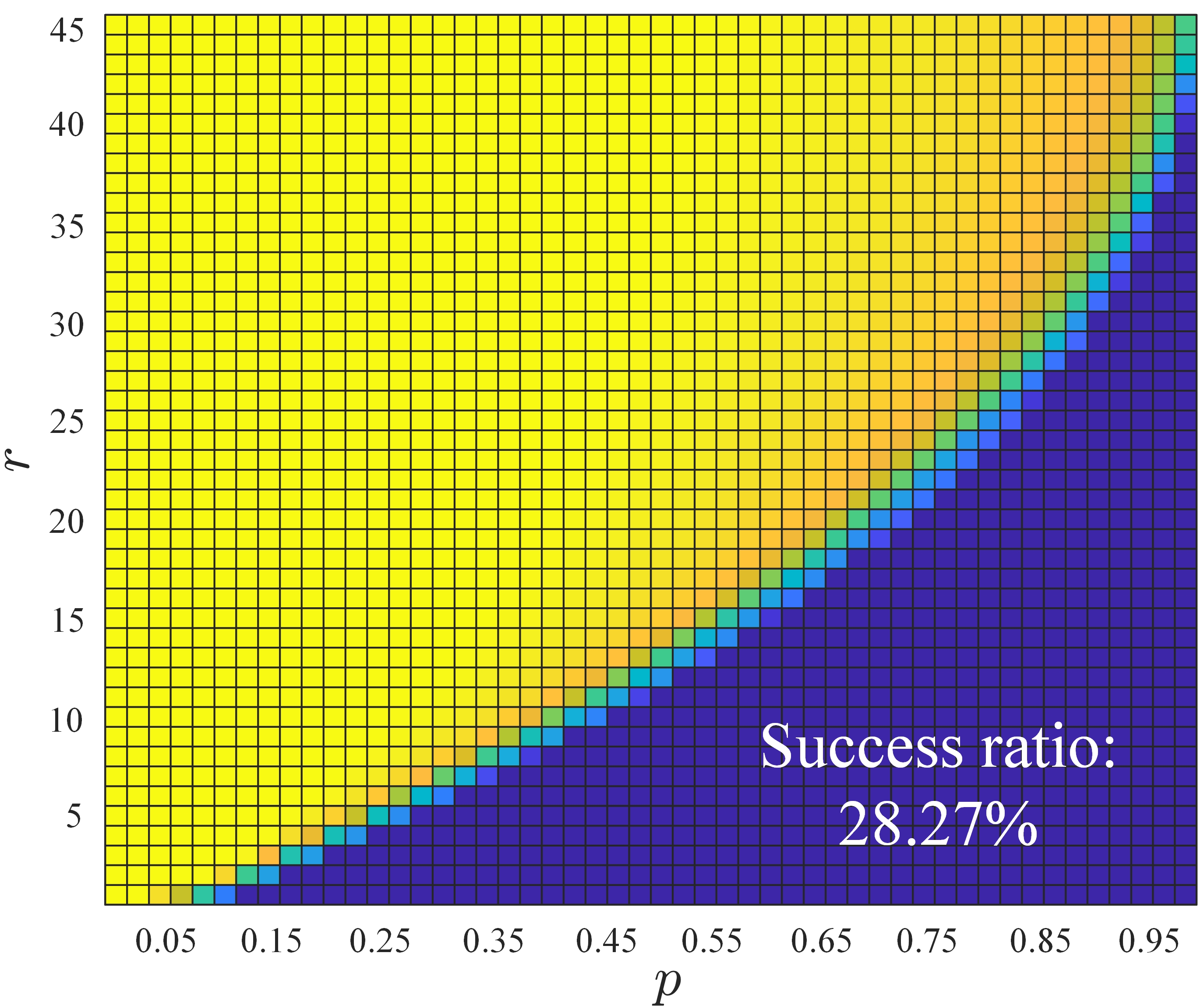}}\vskip 0pt
			\centerline{\scriptsize {(c). GTNN-HOW}}\vskip -3pt
			\centerline{ }
		\end{minipage}\hspace{23mm}
		\begin{minipage}{0.23\linewidth}
			\footnotesize
			\vspace{1pt}
			\centerline{\includegraphics[width=4.7cm]{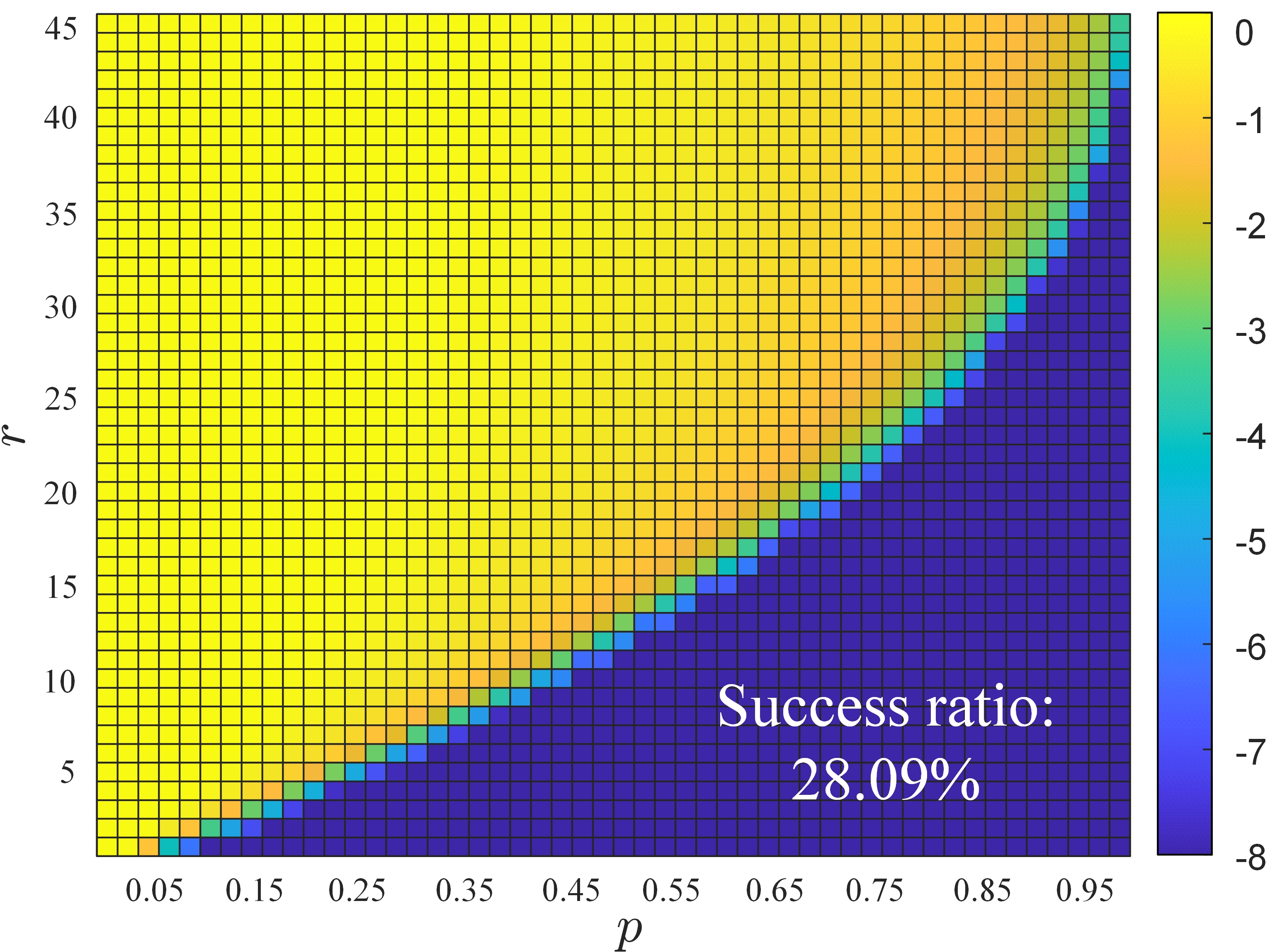}}\vskip 0pt
			\centerline{\scriptsize {(d). GTNN-HOC}}\vskip -3pt
			\centerline{ }
		\end{minipage}
		\caption{Algorithm phase transition diagrams versus rank and SRs.}
		\label{Phase_TC}
	\end{figure}
	\begin{figure}[htb]
		\centering
		\begin{minipage}{0.23\linewidth}
			\footnotesize
			\vspace{1pt}
			\centerline{\includegraphics[width=4.2cm]{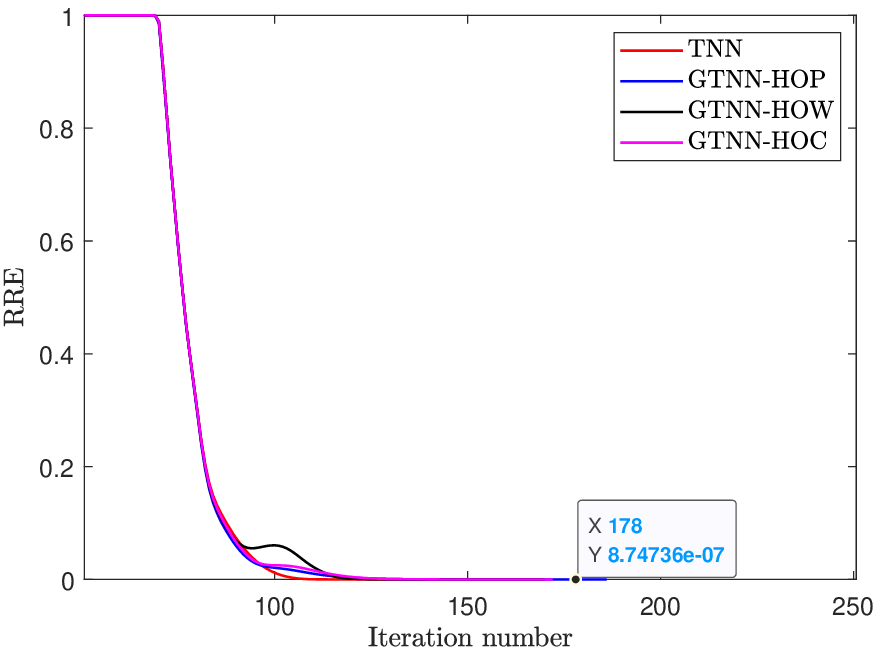}}\vskip 0pt
			\centerline{\scriptsize {(a)}}\vskip -3pt
			\centerline{ }
		\end{minipage}\hspace{23mm}
		\begin{minipage}{0.23\linewidth}
			\footnotesize
			\vspace{1pt}
			\centerline{\includegraphics[width=4.2cm]{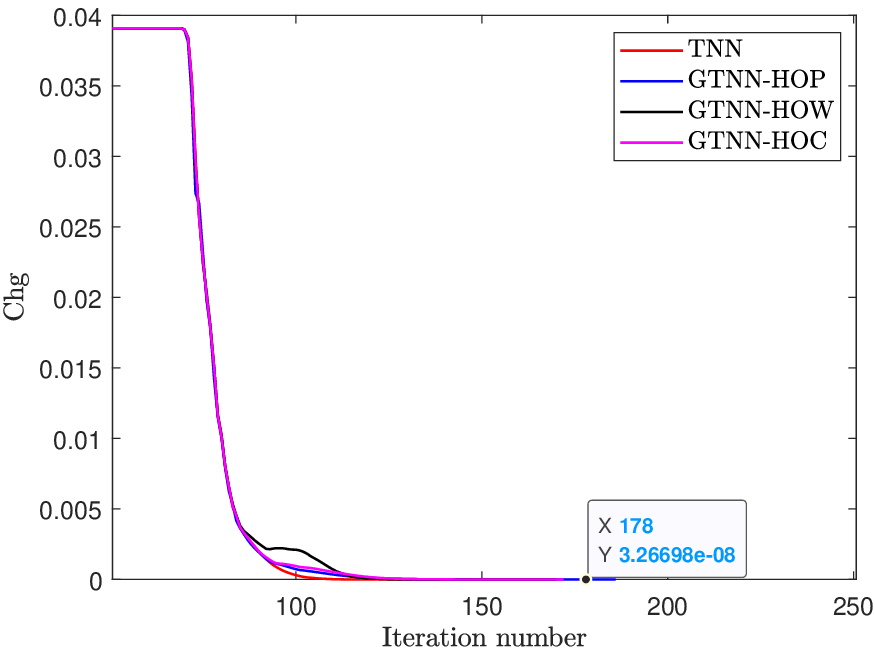}}\vskip 0pt
			\centerline{\scriptsize {(b)}}\vskip -3pt
			\centerline{ }
		\end{minipage}
		\caption{Convergence curves of our algorithms.}
		\label{RRE_convergence}
	\end{figure} 
	\section{EXPERIMENTAL RESULTS}\label{Results}
	In this section, we test the proposed algorithms on synthetic data, real-world images and videos. All simulations are conducted using a computer with 3.0 GHz CPU and 16 GB memory. To evaluate the effectiveness of our algorithms, we compare them with the state-of-art methods, including TNN~\cite{LCYPAMItensor2019}, WTNN~\cite{MUYWTNN2020}, PSTNN~\cite{JiangTXPSTNN2020} and IRTNN~\cite{WangHIRTNN2022}. For the IRTNN, the $\ell_p$-norm with $p=0.5$ is suggested. Besides, for all ADMM-based approaches, the penalty parameter $\rho$ is initialized to $10^{-4}$, and is updated using $\rho_{n+1}=\mu\rho_n$ with $\mu=1.2$. Their termination conditions are set the same, i.e., $I_m = 500$ and $\xi=10^{-4}$.
	Moreover, the recommended setting of the parameters for the competing algorithms is adopted, and we suggest $\sigma = \sqrt{2}\lambda$ and $\gamma = \lambda$ for our methods. Furthermore, $p=0.6$ and $p=0.3$ are used for GTNN-HOP, which are denoted as GTNN-HOP$_{0.6}$ and GTNN-HOP$_{0.3}$, respectively.
	
	\begin{table*}[htb]
		\caption{\small {Real-world image restoration results for the random and fixed masks in terms of PSNR, SSIM, RMSE and runtime. The best and second best results for each row are highlighted in bold and underlined. The results are based on the average of $20$ independent runs.}}  
		\begin{center}
			\setlength{\tabcolsep}{1mm}{
				\begin{tabular}{cccccccccc}
					\hline
					\multicolumn{1}{c}{} &\multicolumn{1}{c}{} &\multicolumn{1}{c}{PSTNN\cite{JiangTXPSTNN2020}} & \multicolumn{1}{c}{IRTNN\cite{WangHIRTNN2022}}  & \multicolumn{1}{c}{WTNN\cite{MUYWTNN2020}}&\multicolumn{1}{c}{TNN\cite{LuCTNN2018}}&\multicolumn{1}{c}{GTNN-HOP$_{0.6}$} & \multicolumn{1}{c}{GTNN-HOP$_{0.3}$}&\multicolumn{1}{c}{GTNN-HOW} & \multicolumn{1}{c}{GTNN-HOC} \\
					\hline
					\multirow{4}{*}{\begin{tabular}[c]{@{}c@{}}SR = $20\%$\end{tabular}} & \multicolumn{1}{c}{PSNR} &\multicolumn{1}{c}{15.864} & \multicolumn{1}{c}{ 22.972} &\multicolumn{1}{c}{{  {25.069}}} & \multicolumn{1}{c}{ 25.434} &\multicolumn{1}{c}{ \bf 26.675}&\multicolumn{1}{c}{ 26.479}&\multicolumn{1}{c}{24.203} & \multicolumn{1}{c}{\underline{26.522}} \\
					& \multicolumn{1}{c}{SSIM} &\multicolumn{1}{c}{0.1747} & \multicolumn{1}{c}{ 0.3726} &\multicolumn{1}{c}{{  {0.5413}}} & \multicolumn{1}{c}{ 0.5935} &\multicolumn{1}{c}{ \bf 0.6472}&\multicolumn{1}{c}{ 0.6336} &\multicolumn{1}{c}{0.5162} & \multicolumn{1}{c}{ \underline{0.6406}}\\
					& \multicolumn{1}{c}{RMSE} &\multicolumn{1}{c}{0.1749} & \multicolumn{1}{c}{ 0.0733} &\multicolumn{1}{c}{{  {0.0577}}} & \multicolumn{1}{c}{ 0.0553} &\multicolumn{1}{c}{ \bf 0.0481}&\multicolumn{1}{c}{ 0.0492} &\multicolumn{1}{c}{0.0639} & \multicolumn{1}{c}{ \underline{0.0489}} \\
					& \multicolumn{1}{c}{Runtime} &\multicolumn{1}{c}{8.3445} & \multicolumn{1}{c}{  64.675} &\multicolumn{1}{c}{{ {2.6020}}} & \multicolumn{1}{c}{ \bf 1.7222} &\multicolumn{1}{c}{ 2.7842}&\multicolumn{1}{c}{ {2.8337}} &\multicolumn{1}{c}{\underline{2.5909}} & \multicolumn{1}{c}{ 2.7725}\\
					\hline
					\multirow{4}{*}{\begin{tabular}[c]{@{}c@{}}SR = $40\%$\end{tabular}} & \multicolumn{1}{c}{PSNR} &\multicolumn{1}{c}{28.310} & \multicolumn{1}{c}{ 30.747} &\multicolumn{1}{c}{{ {30.981}}} & \multicolumn{1}{c}{ 30.869} &\multicolumn{1}{c}{ 32.927}&\multicolumn{1}{c}{\underline {33.175}}&\multicolumn{1}{c}{32.025} & \multicolumn{1}{c}{\bf 33.217} \\
					& \multicolumn{1}{c}{SSIM} &\multicolumn{1}{c}{0.7018} & \multicolumn{1}{c}{ 0.7460} &\multicolumn{1}{c}{{  {0.8018}}} & \multicolumn{1}{c}{ 0.8319} &\multicolumn{1}{c}{ 0.8816}&\multicolumn{1}{c}{ \underline{0.8833}} &\multicolumn{1}{c}{0.8439} & \multicolumn{1}{c}{ \bf 0.8860}\\
					& \multicolumn{1}{c}{RMSE} &\multicolumn{1}{c}{0.0411} & \multicolumn{1}{c}{ 0.0307} &\multicolumn{1}{c}{{  {0.0296}}} & \multicolumn{1}{c}{ 0.0299} &\multicolumn{1}{c}{ 0.0237}&\multicolumn{1}{c}{ \underline{0.0230}} &\multicolumn{1}{c}{0.0263} & \multicolumn{1}{c}{\bf 0.0229} \\
					& \multicolumn{1}{c}{Runtime} &\multicolumn{1}{c}{8.3004} & \multicolumn{1}{c}{ 31.919} &\multicolumn{1}{c}{{ {2.8260}}} & \multicolumn{1}{c}{ \bf 1.7146} &\multicolumn{1}{c}{ 2.2238}&\multicolumn{1}{c}{ {2.3617}} &\multicolumn{1}{c}{2.4715} & \multicolumn{1}{c}{ \underline{2.2028}}\\
					\hline
					\multirow{4}{*}{\begin{tabular}[c]{@{}c@{}}SR = $60\%$\end{tabular}} & \multicolumn{1}{c}{PSNR} &\multicolumn{1}{c}{37.336} & \multicolumn{1}{c}{ 38.875} &\multicolumn{1}{c}{{ {37.692}}} & \multicolumn{1}{c}{ 36.813} &\multicolumn{1}{c}{ 39.284}&\multicolumn{1}{c}{ \underline{39.668}}&\multicolumn{1}{c}{39.164} & \multicolumn{1}{c}{\bf 39.737} \\
					& \multicolumn{1}{c}{SSIM} &\multicolumn{1}{c}{0.9163} & \multicolumn{1}{c}{ 0.9295} &\multicolumn{1}{c}{{  {0.9401}}} & \multicolumn{1}{c}{ 0.9420} &\multicolumn{1}{c}{ 0.9640}&\multicolumn{1}{c}{ \underline{0.9647}} &\multicolumn{1}{c}{0.9554} & \multicolumn{1}{c}{ \bf 0.9655}\\
					& \multicolumn{1}{c}{RMSE} &\multicolumn{1}{c}{0.0145} & \multicolumn{1}{c}{ 0.0122} &\multicolumn{1}{c}{{  {0.0137}}} & \multicolumn{1}{c}{ 0.0151} &\multicolumn{1}{c}{ 0.0114}&\multicolumn{1}{c}{ \underline{0.0109}} &\multicolumn{1}{c}{0.0116} & \multicolumn{1}{c}{ \bf 0.0108} \\
					& \multicolumn{1}{c}{Runtime} &\multicolumn{1}{c}{8.1247} & \multicolumn{1}{c}{  22.688} &\multicolumn{1}{c}{{ {2.9120}}} & \multicolumn{1}{c}{\bf 1.7745} &\multicolumn{1}{c}{ 2.2070}&\multicolumn{1}{c}{ {2.2249}} &\multicolumn{1}{c}{2.2513} & \multicolumn{1}{c}{ \underline{2.1939}}\\
					\hline
					\multirow{4}{*}{\begin{tabular}[c]{@{}c@{}}SR = $80\%$\end{tabular}} & \multicolumn{1}{c}{PSNR} &\multicolumn{1}{c}{46.503} & \multicolumn{1}{c}{ \bf 47.701} &\multicolumn{1}{c}{{ {45.868}}} & \multicolumn{1}{c}{ 44.544} &\multicolumn{1}{c}{ 47.119}&\multicolumn{1}{c}{ {47.602}}&\multicolumn{1}{c}{47.461} & \multicolumn{1}{c}{ \underline{47.678}} \\
					& \multicolumn{1}{c}{SSIM} &\multicolumn{1}{c}{0.9833} & \multicolumn{1}{c}{ 0.9847} &\multicolumn{1}{c}{{  {0.9854}}} & \multicolumn{1}{c}{ 0.9852} &\multicolumn{1}{c}{ 0.9904}&\multicolumn{1}{c}{ \underline{0.9907}} &\multicolumn{1}{c}{0.9892} & \multicolumn{1}{c}{ \bf 0.9908}\\
					& \multicolumn{1}{c}{RMSE} &\multicolumn{1}{c}{0.0050} & \multicolumn{1}{c}{ 0.0044} &\multicolumn{1}{c}{{  {0.0053}}} & \multicolumn{1}{c}{ 0.0062} &\multicolumn{1}{c}{ 0.0046}&\multicolumn{1}{c}{ \bf{0.0043}} &\multicolumn{1}{c}{{0.0044}} & \multicolumn{1}{c}{ \bf 0.0043} \\
					& \multicolumn{1}{c}{Runtime} &\multicolumn{1}{c}{8.1805} & \multicolumn{1}{c}{  17.932} &\multicolumn{1}{c}{{ {3.1095}}} & \multicolumn{1}{c}{ \bf 1.8165} &\multicolumn{1}{c}{ 2.2006}&\multicolumn{1}{c}{ {2.1973}} &\multicolumn{1}{c}{\underline{2.0453}} & \multicolumn{1}{c}{ 2.1534}\\
					\hline
					\multirow{4}{*}{\begin{tabular}[c]{@{}c@{}}Fixed mask\end{tabular}} & \multicolumn{1}{c}{PSNR} &\multicolumn{1}{c}{20.039} & \multicolumn{1}{c}{ 29.739} &\multicolumn{1}{c}{{ {26.688}}} & \multicolumn{1}{c}{ 30.528} &\multicolumn{1}{c}{ 31.423}&\multicolumn{1}{c}{ {31.707}}&\multicolumn{1}{c}{\underline{31.717}} & \multicolumn{1}{c}{ \bf{31.756}} \\
					& \multicolumn{1}{c}{SSIM} &\multicolumn{1}{c}{0.6442} & \multicolumn{1}{c}{ 0.8307} &\multicolumn{1}{c}{{  {0.7778}}} & \multicolumn{1}{c}{ 0.8750} &\multicolumn{1}{c}{ 0.8992}&\multicolumn{1}{c}{ {0.9066}} &\multicolumn{1}{c}{\underline{0.9073}} & \multicolumn{1}{c}{ \bf 0.9085}\\
					& \multicolumn{1}{c}{RMSE} &\multicolumn{1}{c}{0.1037} & \multicolumn{1}{c}{ 0.0339} &\multicolumn{1}{c}{{  {0.0499}}} & \multicolumn{1}{c}{ 0.0307} &\multicolumn{1}{c}{ 0.0278}&\multicolumn{1}{c}{ {0.0270}} &\multicolumn{1}{c}{\underline{0.0269}} & \multicolumn{1}{c}{ \bf 0.0268} \\
					& \multicolumn{1}{c}{Runtime} &\multicolumn{1}{c}{8.6217} & \multicolumn{1}{c}{  62.100} &\multicolumn{1}{c}{{ {3.9565}}} & \multicolumn{1}{c}{ \bf 1.7405} &\multicolumn{1}{c}{ 2.2687}&\multicolumn{1}{c}{ {2.2209}} &\multicolumn{1}{c}{\underline{2.0300}} & \multicolumn{1}{c}{ 2.1611}\\
					\hline
			\end{tabular}}
			\vspace{-2em}
			\label{SRs_Imageinpainting}
		\end{center}
	\end{table*}

	\subsection{Synthetic Data}	
	We first conduct experiments to verify the superiority of our algorithms over the convex TNN method using synthetic data. A low-rank tensor $\bm{\mathcal{X}} \in \mathbb{R}^{n\times n \times n}$ with tubal rank $r$ is generated by t-product $\bm{\mathcal{X}} = \bm{\mathcal{X}}_1*\bm{\mathcal{X}}_2$ where the entries of $\bm{\mathcal{X}}_1 \in \mathbb{R}^{n\times r \times n}$ and $\bm{\mathcal{X}}_2 \in \mathbb{R}^{r\times n \times n}$ are standard Gaussian distributed. To evaluate the recovery performance, the relative reconstruction error (RRE) defined as
	\begin{equation*}
		{\rm RRE} = \|\bm{\mathcal{M}} - \bm{\mathcal{X}}\|_F/\left\|\bm{\mathcal{X}}\right\|_F
	\end{equation*}
	 where $\bm{\mathcal{M}}$ is the estimated low-rank tensor, is employed. Moreover, the performance of all approaches is evaluated based on the average results of $20$ independent runs.
	
	The incomplete tensor $\bm{\mathcal{X}}_\Omega$ is constructed by sampling $m=pn^3$ entries uniformly from $\bm{\mathcal{X}}$, where $p$ is the sampling rate (SR). The impact of varying tubal rank $r$ and $p$ on the recovery performance is examined. We choose $r=1,2,\cdots, 45$ and $p$ from the set $[0.01:0.02:0.99]$, and consider a trail as success if ${\rm RRE}\leq 10^{-4}$. The results in terms of log-scale $\rm RRE$ are shown in Fig.~\ref{Phase_TC}.
	For each pair $(r,p)$, the blue and yellow region reflects the successful and unsuccessful recovery, respectively. It is observed that our nonconvex surrogates for the rank have a bigger success area than the convex TNN technique.
	Besides, we perform experiments to verify the convergence of our algorithms. Apart from RRE, additional evaluation metrics are introduced:
	\begin{equation*}
		{\rm Chg} = \max\{{\rm Chg}\bm{\mathcal{M}},{\rm Chg}\bm{\mathcal{E}},{\rm Chg}\bm{\mathcal{X}}\}
	\end{equation*}
	where ${\rm Chg}\bm{\mathcal{M}}= \left\|\bm{\mathcal{M}}_{n+1}-\bm{\mathcal{M}}_n\right\|_\infty$, ${\rm Chg}\bm{\mathcal{E}}= \left\|\bm{\mathcal{E}}_{n+1}-\bm{\mathcal{E}}_n\right\|_\infty$ and ${\rm Chg}\bm{\mathcal{X}}=\left\|\bm{\mathcal{X}}-\bm{\mathcal{M}}_{n+1} -\bm{\mathcal{E}}_{n+1}\right\|_\infty$. Fig.~\ref{RRE_convergence} plots the convergence curves of our algorithms and TNN. It is seen that although we adopt nonconvex regularizers to replace the tensor nuclear norm, the developed techniques converge in terms of RRE, ${\rm Chg}\bm{\mathcal{M}}$, ${\rm Chg}\bm{\mathcal{E}}$ and ${\rm Chg}\bm{\mathcal{X}}$, which is consistent with Theorem~\ref{theorem-convergence}.
		
	\subsection{Image Inpainting}
\begin{figure}[htb]
		\centering
		\centerline{\scriptsize {Image-1~~~Image-2~~~Image-3~~~Image-4~~~Image-5~~~Image-6~~~Image-7~~~Image-8}}\vskip 1pt
		\includegraphics[width=8.5cm]{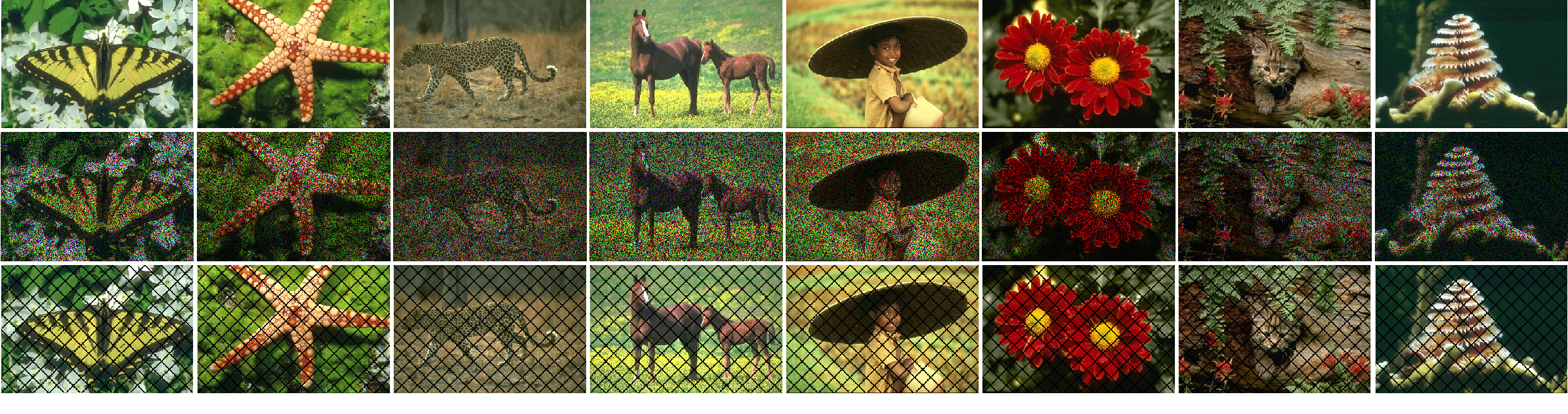}
		\vspace{-0.5em}
		\caption{Test images}\label{TC_8images}
	\end{figure}

	For real-world data, we first evaluate all algorithms in the task of natural image inpainting since color images comprise three (R, G and B) channels and can be approximated as a $3$rd-order low-rank-tubal tensor~\cite{WangHIRTNN2022}. Eight images from the Berkeley Segmentation Database (BSD)~\cite{BSDdataset} are used and two types of masks, i.e., random and fixed masks, are investigated. Fig.~\ref{TC_8images} shows the adopted color images where the
	first, second and third rows are the original images, incomplete images covered by a random mask and incomplete images covered by a fixed mask, respectively. The random mask means that the missing entries are randomly selected, while the fixed mask contains regular stripes. For the random mask, the impact of SR on recovery performance is investigated and we consider different SRs ($20\%$, $40\%$, $60\%$, and $80\%$) for each image. The performance metrics
	are evaluated by peak signal-to-noise (PSNR), structural similarity index (SSIM) and root mean square error (RMSE), which are defined as:
	\begin{equation*}
		\begin{split}
			{\rm PSNR}(\bm{\mathcal{M}},\bm{\mathcal{X}})&=\frac{1}{n_3}\sum_{j=1}^{n_3}{\rm PSNR}(\bm{\mathcal{M}}^{(j)},\bm{\mathcal{X}}^{(j)})\\
			{\rm SSIM}(\bm{\mathcal{M}},\bm{\mathcal{X}})&=\frac{1}{n_3}\sum_{j=1}^{n_3}{\rm SSIM}(\bm{\mathcal{M}}^{(j)},\bm{\mathcal{X}}^{(j)})\\
			{\rm RMSE}(\bm{\mathcal{M}},\bm{\mathcal{X}})&=\frac{1}{n_3}\sum_{j=1}^{n_3}\frac{\|\bm{\mathcal{M}}^{(j)}-\bm{\mathcal{X}}^{(j)}\|_F}{\sqrt{n_1\times n_2}}
		\end{split}
	\end{equation*}

	\begin{figure*}[htb]
		\centering
		
		\begin{minipage}{0.0915\linewidth}
			
			\footnotesize
			\vspace{1pt}
			\centerline{\includegraphics[width=3.6cm]{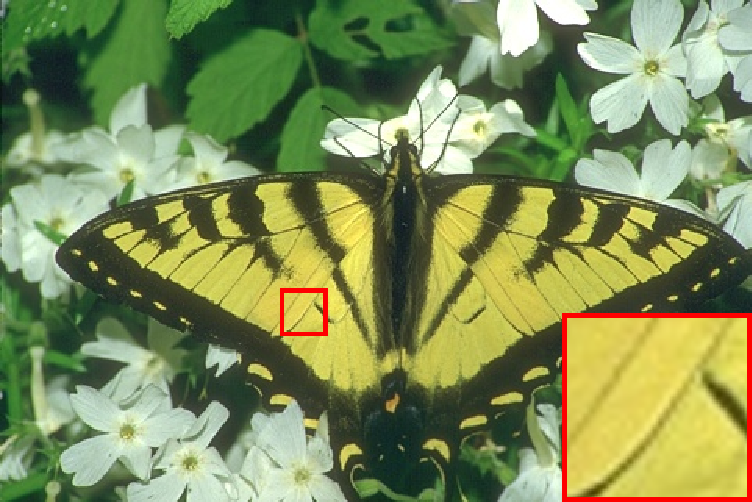}}\vskip 0pt
			\centerline{\scriptsize {Original image}}\vskip -3pt
		\end{minipage}\hspace{19mm}
		\begin{minipage}{0.0915\linewidth}
			\footnotesize
			\vspace{1pt}
			\centerline{\includegraphics[width=3.6cm]{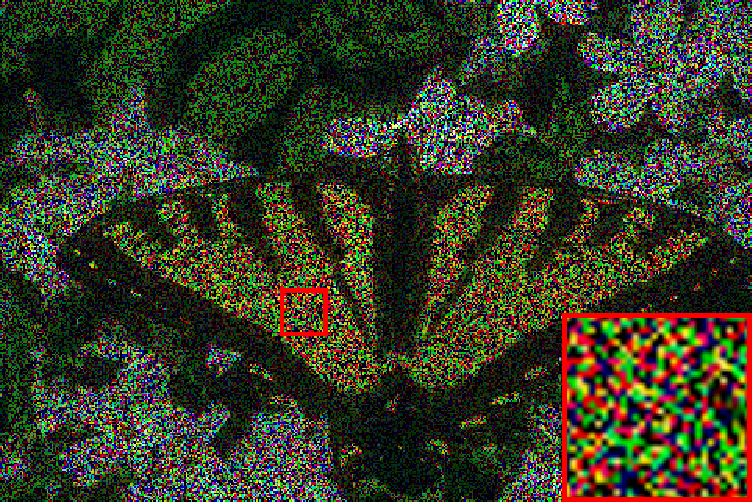}}\vskip 0pt
			\centerline{\scriptsize {Observed image (PSNR, SSIM, RMSE)}}\vskip -3pt
		\end{minipage}\hspace{19mm}
		\begin{minipage}{0.0915\linewidth}
			\footnotesize
			\vspace{1pt}
			\centerline{\includegraphics[width=3.6cm]{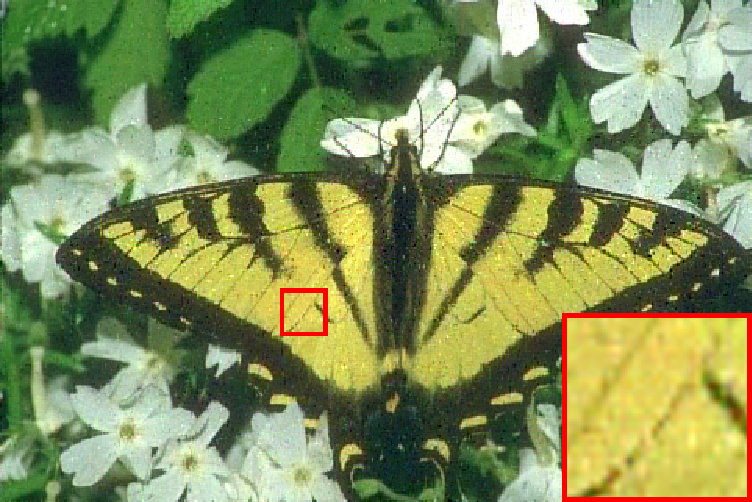}}\vskip 0pt
			\centerline{\scriptsize {{PSTNN (27.32, 0.662, 0.043)}}}\vskip -3pt
		\end{minipage}\hspace{19mm}
		\begin{minipage}{0.0915\linewidth}
			\footnotesize
			\vspace{1pt}
			\centerline{\includegraphics[width=3.6cm]{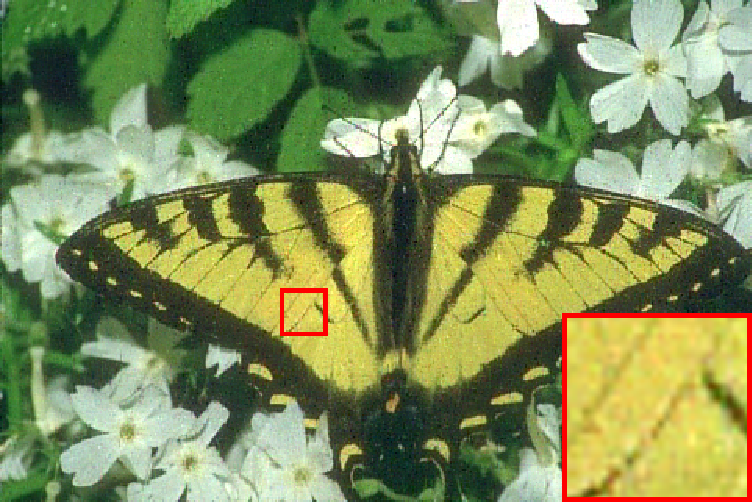}}\vskip 0pt
			\centerline{\scriptsize {IRTNN (27.95, 0.672, 0.040)}}\vskip -3pt
		\end{minipage}\hspace{19mm}
		\begin{minipage}{0.0915\linewidth}
			\footnotesize
			\vspace{1pt}
			\centerline{\includegraphics[width=3.6cm]{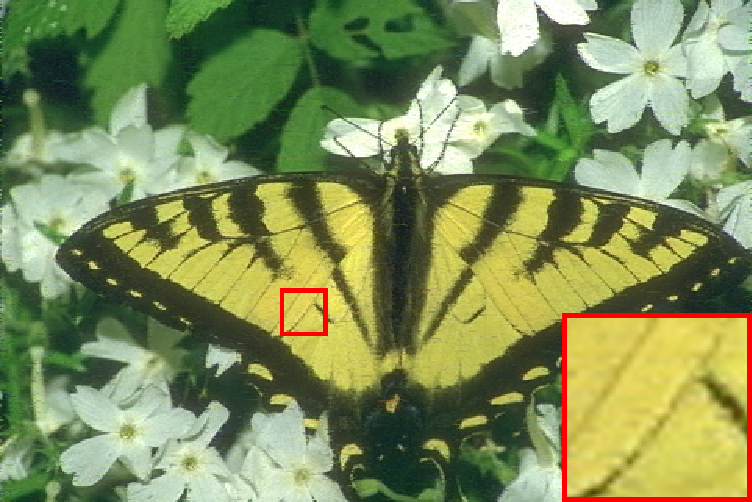}}\vskip 0pt
			\centerline{\scriptsize {WTNN (29.95, 0.815, 0.032)}}\vskip -3pt
		\end{minipage}
		\vskip 3pt
		\begin{minipage}{0.0915\linewidth}
			\footnotesize
			\vspace{1pt}
			\centerline{\includegraphics[width=3.6cm]{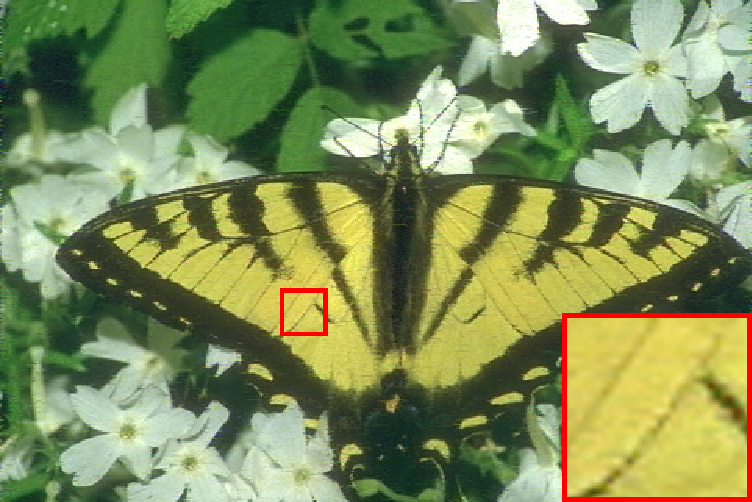}}\vskip 0pt
			\centerline{\scriptsize {TNN (29.16, 0.803, 0.035)}}\vskip -3pt
		\end{minipage}\hspace{19mm}
		\begin{minipage}{0.0915\linewidth}
			\footnotesize
			\vspace{1pt}
			\centerline{\includegraphics[width=3.6cm]{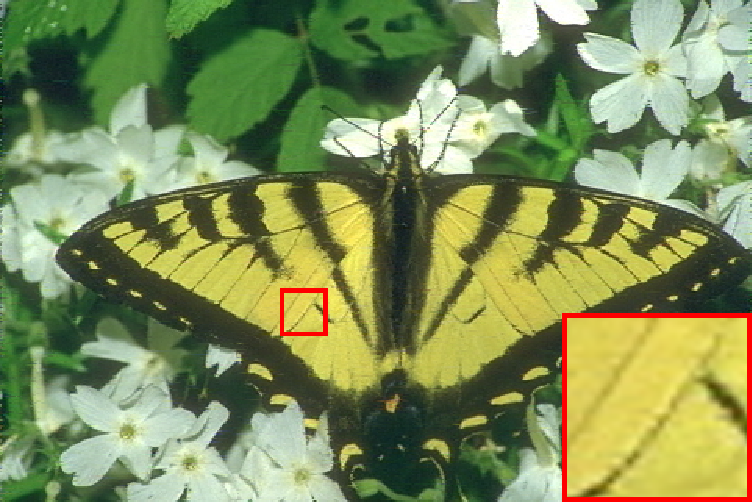}}\vskip 0pt
			\centerline{\scriptsize {GTNN-HOP$_{0.6}$(31.41, 0.869, 0.027)}}\vskip -3pt
		\end{minipage}\hspace{19mm}
		\begin{minipage}{0.0915\linewidth}
			\footnotesize
			\vspace{1pt}
			\centerline{\includegraphics[width=3.6cm]{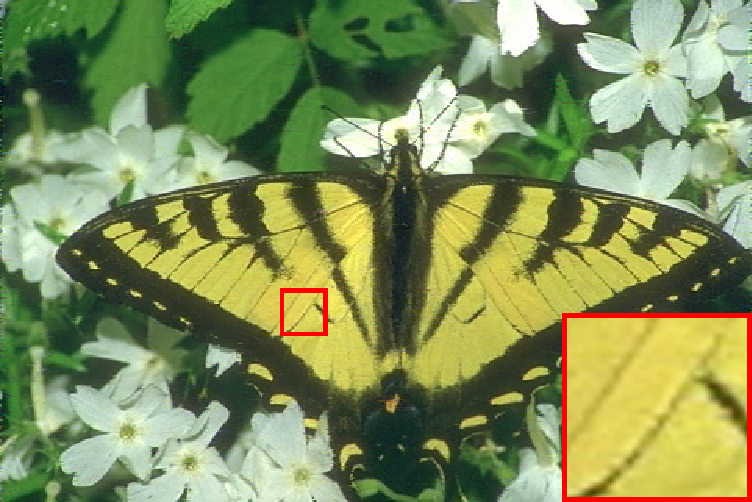}}\vskip 0pt
			\centerline{\scriptsize {GTNN-HOP$_{0.3}$(31.70, 0.872, 0.026)}}\vskip -3pt
		\end{minipage}\hspace{19mm}
		\begin{minipage}{0.0915\linewidth}
			\footnotesize
			\vspace{1pt}
			\centerline{\includegraphics[width=3.6cm]{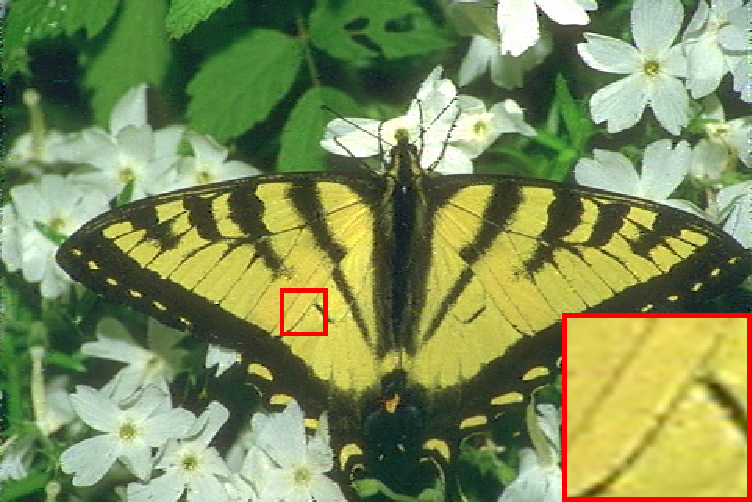}}\vskip 0pt
			\centerline{\scriptsize {GTNN-HOW (30.56, 0.824, 0.030)}}\vskip -3pt
		\end{minipage}\hspace{19mm}
		\begin{minipage}{0.0915\linewidth}
			\footnotesize
			\vspace{1pt}
			\centerline{\includegraphics[width=3.6cm]{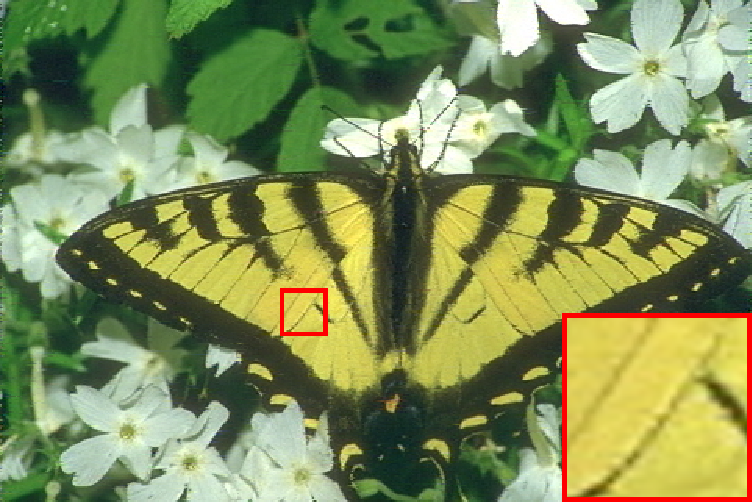}}\vskip 0pt
			\centerline{\scriptsize {GTNN-HOC (31.73, 0.874, 0.026)}}\vskip -3pt
		\end{minipage}
		\caption{Image recovery results for random mask with ${\rm SR}=40\%$ by different algorithms}
		\label{Butterfly_randommask}
	\end{figure*}

	\begin{figure*}[htb]
		\centering
		
		\begin{minipage}{0.0915\linewidth}
			
			\footnotesize
			\vspace{1pt}
			\centerline{\includegraphics[width=3.6cm]{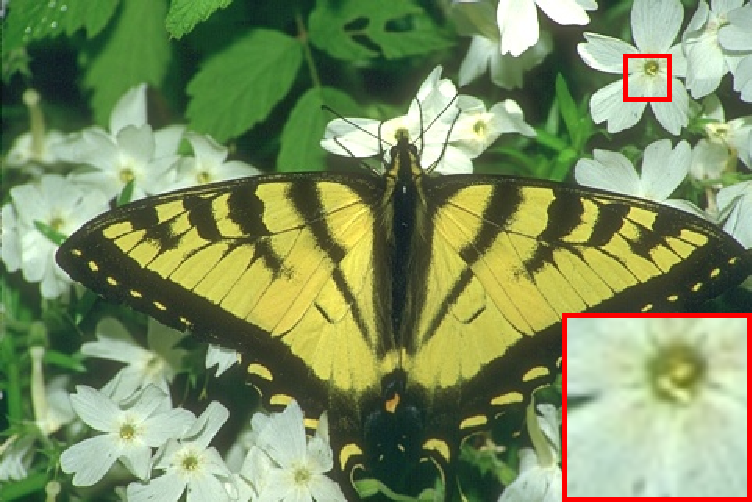}}\vskip 0pt
			\centerline{\scriptsize {Original image}}\vskip -3pt
		\end{minipage}\hspace{19mm}
		\begin{minipage}{0.0915\linewidth}
			\footnotesize
			\vspace{1pt}
			\centerline{\includegraphics[width=3.6cm]{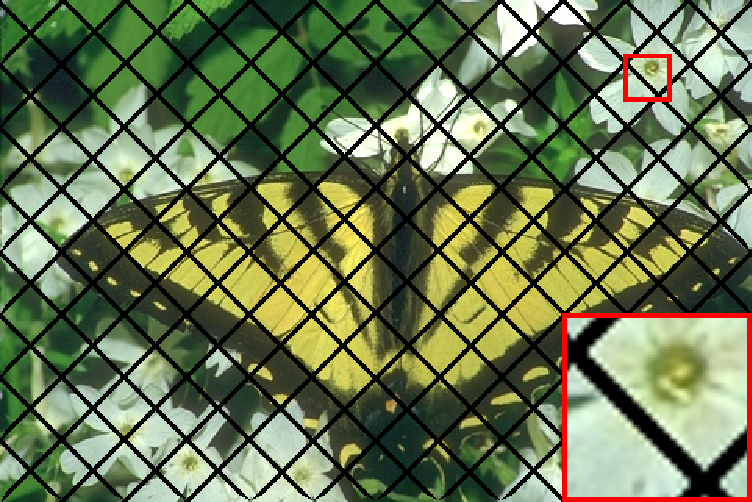}}\vskip 0pt
			\centerline{\scriptsize {Observed image (PSNR, SSIM, RMSE)}}\vskip -3pt
		\end{minipage}\hspace{19mm}
		\begin{minipage}{0.0915\linewidth}
			\footnotesize
			\vspace{1pt}
			\centerline{\includegraphics[width=3.6cm]{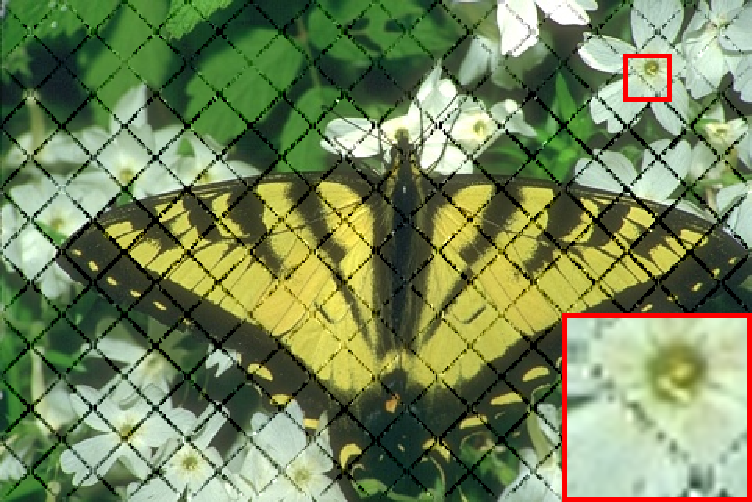}}\vskip 0pt
			\centerline{\scriptsize {{PSTNN (18.23, 0.618, 0.123)}}}\vskip -3pt
		\end{minipage}\hspace{19mm}
		\begin{minipage}{0.0915\linewidth}
			\footnotesize
			\vspace{1pt}
			\centerline{\includegraphics[width=3.6cm]{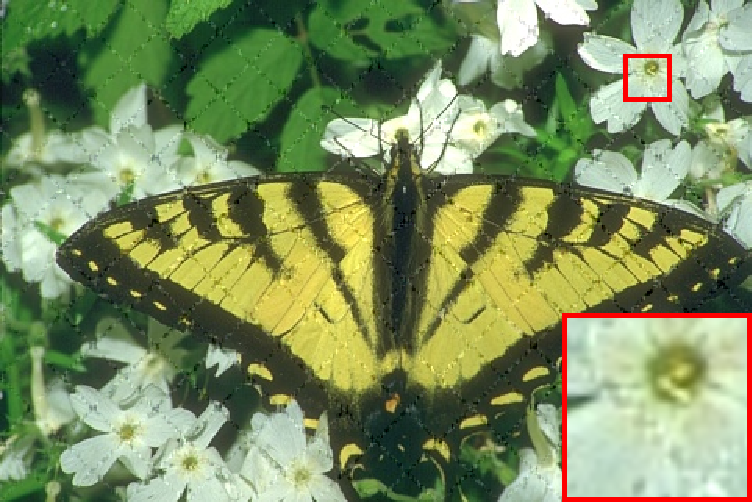}}\vskip 0pt
			\centerline{\scriptsize {IRTNN (27.20, 0.783, 0.044)}}\vskip -3pt
		\end{minipage}\hspace{19mm}
		\begin{minipage}{0.0915\linewidth}
			\footnotesize
			\vspace{1pt}
			\centerline{\includegraphics[width=3.6cm]{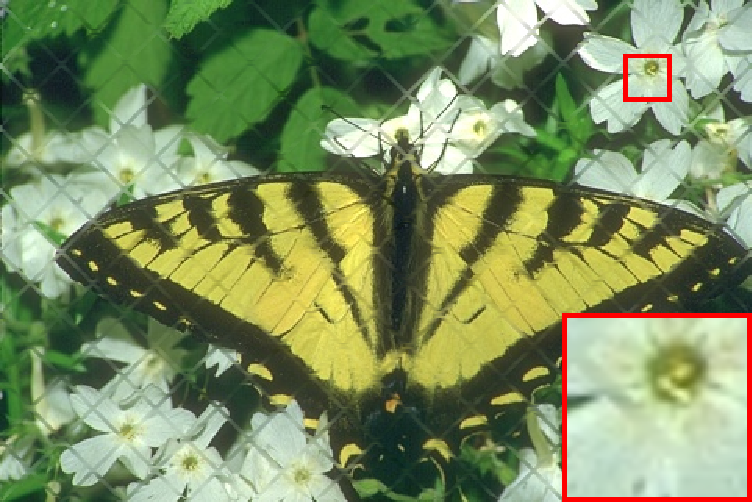}}\vskip 0pt
			\centerline{\scriptsize {WTNN (27.26, 0.823, 0.043)}}\vskip -3pt
		\end{minipage}
		\vskip 3pt
		\begin{minipage}{0.0915\linewidth}
			\footnotesize
			\vspace{1pt}
			\centerline{\includegraphics[width=3.6cm]{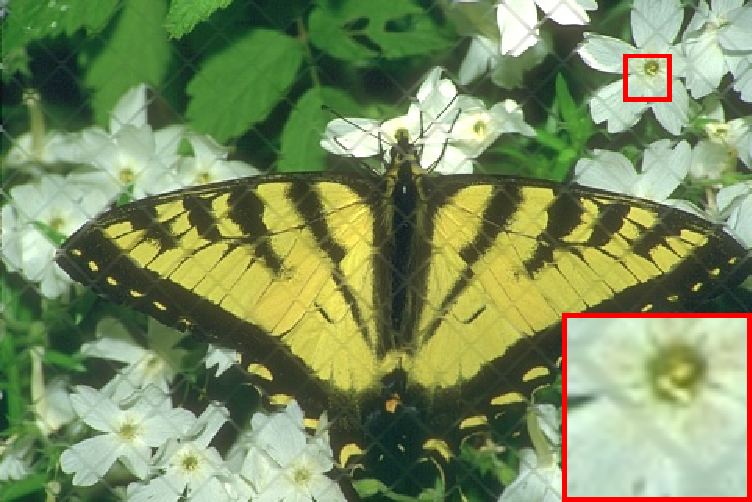}}\vskip 0pt
			\centerline{\scriptsize {TNN (28.31, 0.848, 0.039)}}\vskip -3pt
		\end{minipage}\hspace{19mm}
		\begin{minipage}{0.0915\linewidth}
			\footnotesize
			\vspace{1pt}
			\centerline{\includegraphics[width=3.6cm]{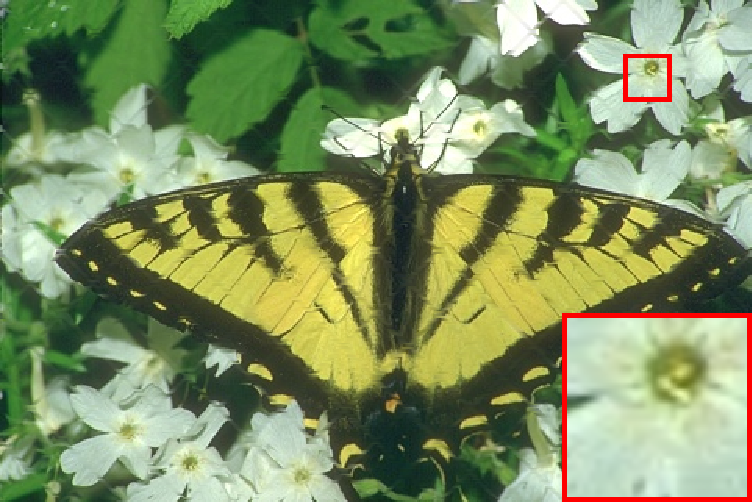}}\vskip 0pt
			\centerline{\scriptsize {GTNN-HOP$_{0.6}$(29.42, 0.885, 0.034)}}\vskip -3pt
		\end{minipage}\hspace{19mm}
		\begin{minipage}{0.0915\linewidth}
			\footnotesize
			\vspace{1pt}
			\centerline{\includegraphics[width=3.6cm]{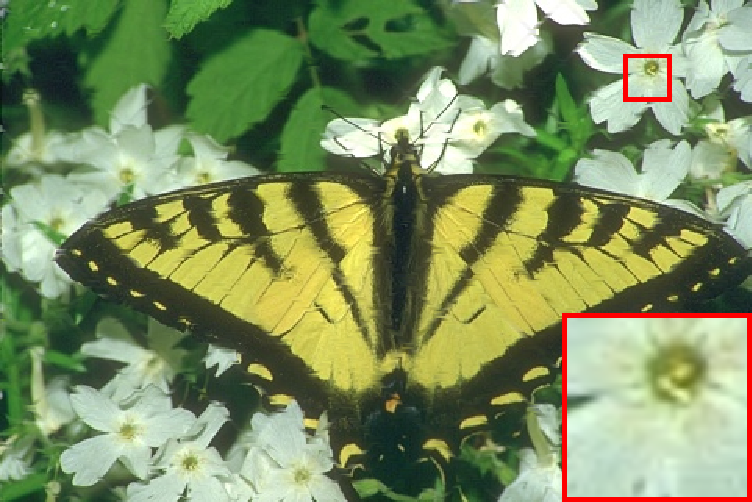}}\vskip 0pt
			\centerline{\scriptsize {GTNN-HOP$_{0.3}$(29.76, 0.896, 0.033)}}\vskip -3pt
		\end{minipage}\hspace{19mm}
		\begin{minipage}{0.0915\linewidth}
			\footnotesize
			\vspace{1pt}
			\centerline{\includegraphics[width=3.6cm]{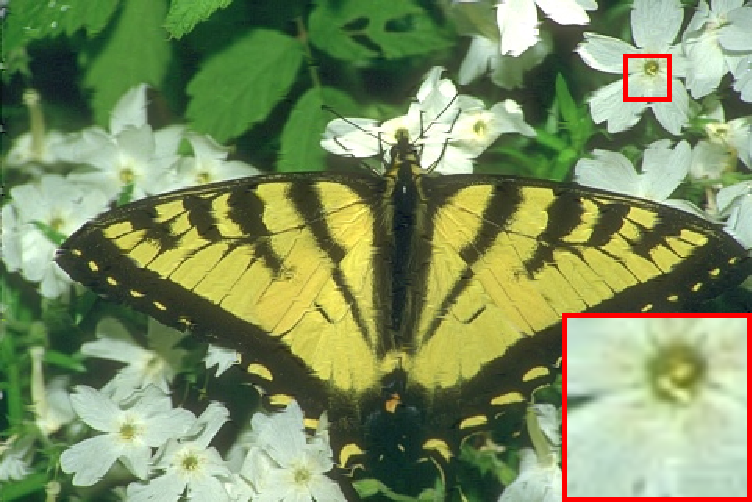}}\vskip 0pt
			\centerline{\scriptsize {GTNN-HOW (29.76, 0.891, 0.033)}}\vskip -3pt
		\end{minipage}\hspace{19mm}
		\begin{minipage}{0.0915\linewidth}
			\footnotesize
			\vspace{1pt}
			\centerline{\includegraphics[width=3.6cm]{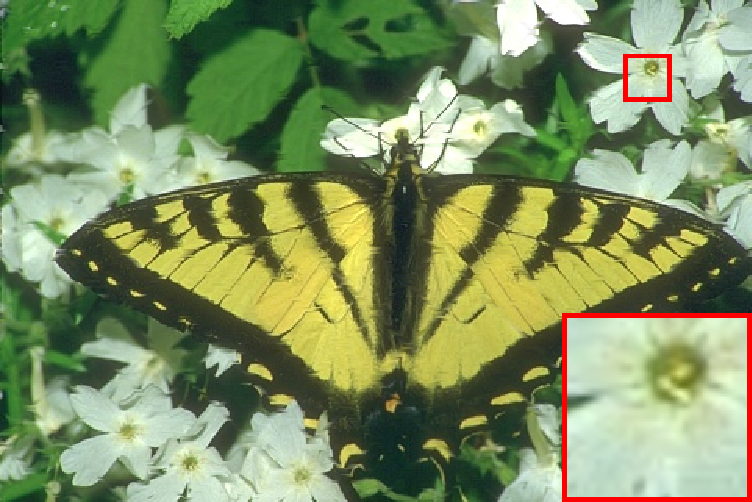}}\vskip 0pt
			\centerline{\scriptsize {GTNN-HOC (29.82, 0.898, 0.032)}}\vskip -3pt
		\end{minipage}
		\caption{Image recovery results for fixed mask by different algorithms}
		\label{Butterfly_fixedmask}
	\end{figure*}
	Table~\ref{SRs_Imageinpainting} tabulates the average recovery results of the eight images for the two masks. It is seen that our proposed algorithms achieve better recovery for most cases in terms of PSNR, SSIM and RMSE. GTNN-HOC yields the best restoration results except for ${\rm SR}=20\%$. Compared with GTNN-HOP$_{0.3}$ which has outstanding performance in random mask, GTNN-HOW can deal with fixed mask well. In addition, compared with IRTNN that requires iterations to find its proximity operator, TNN and our algorithms need less runtimes because they have the closed-form thresholding operators. As the thresholding operator for TNN has a simpler expression than those of our methods, it involves the minimum runtime. 
	To provide visual comparison, Figs.~\ref{Butterfly_randommask} and \ref{Butterfly_fixedmask} show the recovered images of Image-1 for the two masks. The former corresponds to the random mask with ${\rm SR}=40\%$, and it is seen that our methods give clearer images than the remaining approaches. From the restoration results for the fixed mask in Fig.~\ref{Butterfly_fixedmask}, we observe that there are still some apparent stripes in the recovered images generated by PSTNN, IRTNN, WTNN and TNN.

	\begin{figure*}[htb]
		\centering
		\begin{minipage}{0.0915\linewidth}
			\caption*{Original}
			\vspace{-6pt}
			\footnotesize
			\centerline{\includegraphics[width=1.95cm]{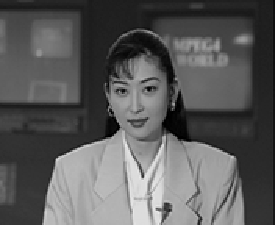}}\vskip 0pt
			\centerline{}\vskip -3pt
		\end{minipage}\hspace{2mm}
		\begin{minipage}{0.0915\linewidth}
			\caption*{Observed}
			\vspace{-5pt}
			\footnotesize
			\centerline{\includegraphics[width=1.95cm]{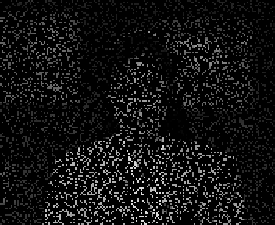}}\vskip 0pt
			\centerline{\scriptsize {(PSNR, SSIM)}}\vskip -3pt
		\end{minipage}\hspace{2mm}
		\begin{minipage}{0.0915\linewidth}
			\footnotesize
			\caption*{PSTNN}
			\vspace{-5pt}
			\centerline{\includegraphics[width=1.95cm]{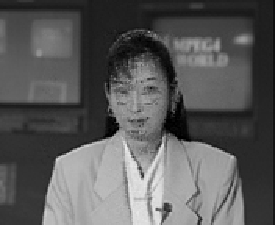}}\vskip 0pt
			\centerline{\scriptsize {{(30.74, 0.9464)}}}\vskip -3pt
		\end{minipage}\hspace{2mm}
		\begin{minipage}{0.0915\linewidth}
			\footnotesize
			\caption*{IRTNN}
			\vspace{-5pt}
			\centerline{\includegraphics[width=1.95cm]{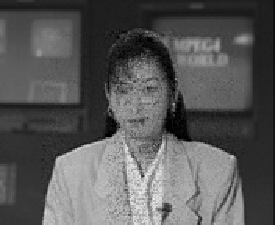}}\vskip 0pt
			\centerline{\scriptsize {(27.56, 0.8646)}}\vskip -3pt
		\end{minipage}\hspace{2mm}
		\begin{minipage}{0.0915\linewidth}
			\footnotesize
			\caption*{WTNN}
			\vspace{-5pt}
			\centerline{\includegraphics[width=1.95cm]{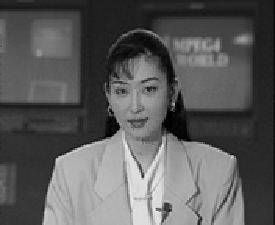}}\vskip 0pt
			\centerline{\scriptsize {(39.79, 0.9853)}}\vskip -3pt
		\end{minipage}\hspace{2mm}
		\begin{minipage}{0.0915\linewidth}
			\footnotesize
			\caption*{TNN}
			\vspace{-5pt}
			\centerline{\includegraphics[width=1.95cm]{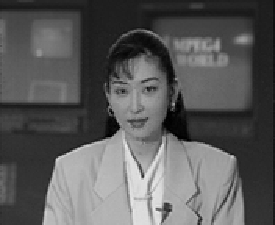}}\vskip 0pt
			\centerline{\scriptsize {(38.66, 0.9819)}}\vskip -3pt
		\end{minipage}\hspace{2mm}
		\begin{minipage}{0.0915\linewidth}
			\footnotesize
			\caption*{GTNN-HOP$_{0.6}$}
			\vspace{-5pt}
			\centerline{\includegraphics[width=1.95cm]{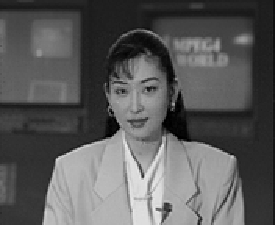}}\vskip 0pt
			\centerline{\scriptsize {(40.87, 0.9857)}}\vskip -3pt
		\end{minipage}\hspace{2mm}
		\begin{minipage}{0.0915\linewidth}
			\footnotesize
			\caption*{GTNN-HOP$_{0.3}$}
			\vspace{-5pt}
			\centerline{\includegraphics[width=1.95cm]{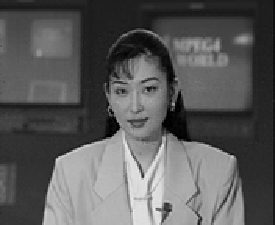}}\vskip 0pt
			\centerline{\scriptsize {(40.72, 0.9834)}}\vskip -3pt
		\end{minipage}\hspace{2mm}
		\begin{minipage}{0.0915\linewidth}
			\footnotesize
			\caption*{GTNN-HOC}
			\vspace{-5pt}
			\centerline{\includegraphics[width=1.95cm]{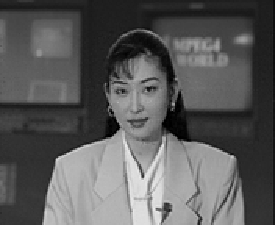}}\vskip 0pt
			\centerline{\scriptsize {(40.81, 0.9845)}}\vskip -3pt
		\end{minipage}	
	
		\begin{minipage}{0.0915\linewidth}
			\vspace{0pt}
			\footnotesize
			\centerline{\includegraphics[width=1.95cm]{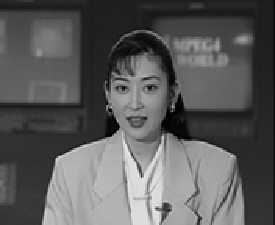}}\vskip 0pt
			\centerline{}\vskip -3pt
		\end{minipage}\hspace{2mm}
		\begin{minipage}{0.0915\linewidth}
			\vspace{1pt}
			\footnotesize
			\centerline{\includegraphics[width=1.95cm]{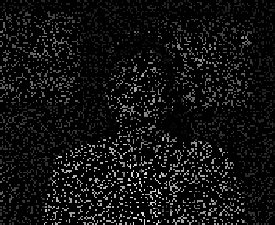}}\vskip 0pt
			\centerline{\scriptsize {(PSNR, SSIM)}}\vskip -3pt
		\end{minipage}\hspace{2mm}
		\begin{minipage}{0.0915\linewidth}
			\footnotesize
			\vspace{1pt}
			\centerline{\includegraphics[width=1.95cm]{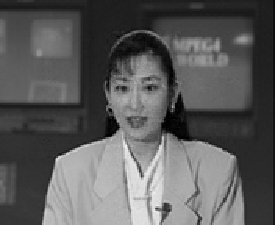}}\vskip 0pt
			\centerline{\scriptsize {{(35.22, 0.9739)}}}\vskip -3pt
		\end{minipage}\hspace{2mm}
		\begin{minipage}{0.0915\linewidth}
			\footnotesize
			\vspace{1pt}
			\centerline{\includegraphics[width=1.95cm]{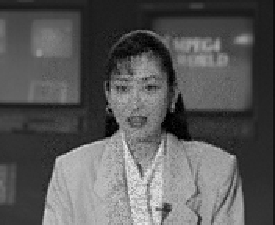}}\vskip 0pt
			\centerline{\scriptsize {(29.04, 0.8827)}}\vskip -3pt
		\end{minipage}\hspace{2mm}
		\begin{minipage}{0.0915\linewidth}
			\footnotesize
			\vspace{1pt}
			\centerline{\includegraphics[width=1.95cm]{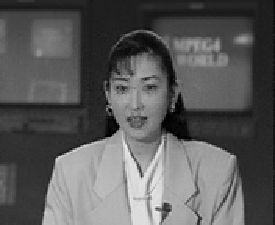}}\vskip 0pt
			\centerline{\scriptsize {(36.92, 0.9731)}}\vskip -3pt
		\end{minipage}\hspace{2mm}
		\begin{minipage}{0.0915\linewidth}
			\footnotesize
			\vspace{1pt}
			\centerline{\includegraphics[width=1.95cm]{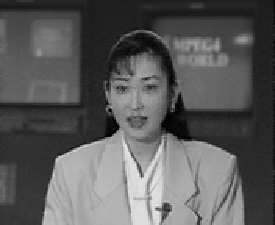}}\vskip 0pt
			\centerline{\scriptsize {(35.87, 0.9681)}}\vskip -3pt
		\end{minipage}\hspace{2mm}
		\begin{minipage}{0.0915\linewidth}
			\footnotesize
			\vspace{1pt}
			\centerline{\includegraphics[width=1.95cm]{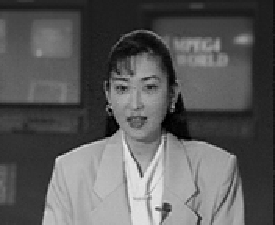}}\vskip 0pt
			\centerline{\scriptsize {(37.78, 0.9739)}}\vskip -3pt
		\end{minipage}\hspace{2mm}
		\begin{minipage}{0.0915\linewidth}
			\footnotesize
			\vspace{1pt}
			\centerline{\includegraphics[width=1.95cm]{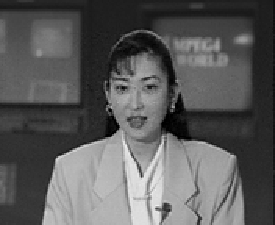}}\vskip 0pt
			\centerline{\scriptsize {(37.56, 0.9694)}}\vskip -3pt
		\end{minipage}\hspace{2mm}
		\begin{minipage}{0.0915\linewidth}
			\footnotesize
			\vspace{1pt}
			\centerline{\includegraphics[width=1.95cm]{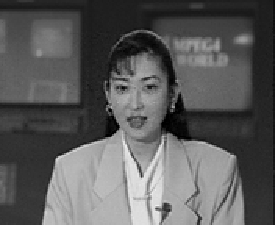}}\vskip 0pt
			\centerline{\scriptsize {(37.68, 0.9713)}}\vskip -3pt
		\end{minipage}	
	
		\begin{minipage}{0.0915\linewidth}
			\vspace{0pt}
			\footnotesize
			\centerline{\includegraphics[width=1.95cm]{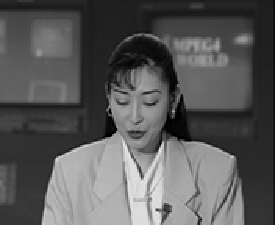}}\vskip 0pt
			\centerline{}\vskip -3pt
		\end{minipage}\hspace{2mm}
		\begin{minipage}{0.0915\linewidth}
			\vspace{1pt}
			\footnotesize
			\centerline{\includegraphics[width=1.95cm]{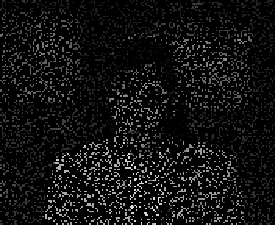}}\vskip 0pt
			\centerline{\scriptsize {(PSNR, SSIM)}}\vskip -3pt
		\end{minipage}\hspace{2mm}
		\begin{minipage}{0.0915\linewidth}
			\footnotesize
			\vspace{1pt}
			\centerline{\includegraphics[width=1.95cm]{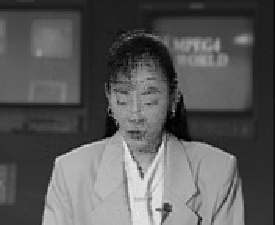}}\vskip 0pt
			\centerline{\scriptsize {{(30.39, 0.9391)}}}\vskip -3pt
		\end{minipage}\hspace{2mm}
		\begin{minipage}{0.0915\linewidth}
			\footnotesize
			\vspace{1pt}
			\centerline{\includegraphics[width=1.95cm]{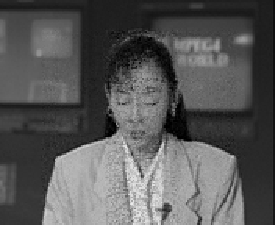}}\vskip 0pt
			\centerline{\scriptsize {(27.64, 0.8600)}}\vskip -3pt
		\end{minipage}\hspace{2mm}
		\begin{minipage}{0.0915\linewidth}
			\footnotesize
			\vspace{1pt}
			\centerline{\includegraphics[width=1.95cm]{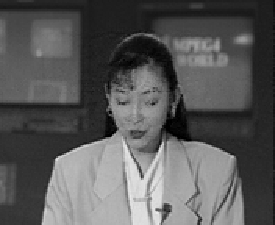}}\vskip 0pt
			\centerline{\scriptsize {(34.38, 0.9561)}}\vskip -3pt
		\end{minipage}\hspace{2mm}
		\begin{minipage}{0.0915\linewidth}
			\footnotesize
			\vspace{1pt}
			\centerline{\includegraphics[width=1.95cm]{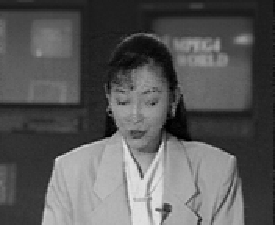}}\vskip 0pt
			\centerline{\scriptsize {(33.63, 0.9499)}}\vskip -3pt
		\end{minipage}\hspace{2mm}
		\begin{minipage}{0.0915\linewidth}
			\footnotesize
			\vspace{1pt}
			\centerline{\includegraphics[width=1.95cm]{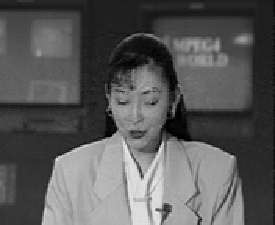}}\vskip 0pt
			\centerline{\scriptsize {(35.08, 0.9573)}}\vskip -3pt
		\end{minipage}\hspace{2mm}
		\begin{minipage}{0.0915\linewidth}
			\footnotesize
			\vspace{1pt}
			\centerline{\includegraphics[width=1.95cm]{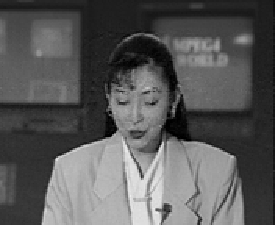}}\vskip 0pt
			\centerline{\scriptsize {(34.83, 0.9509)}}\vskip -3pt
		\end{minipage}\hspace{2mm}
		\begin{minipage}{0.0915\linewidth}
			\footnotesize
			\vspace{1pt}
			\centerline{\includegraphics[width=1.95cm]{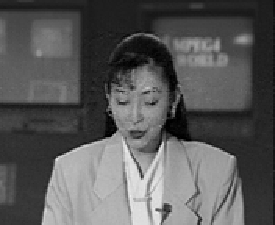}}\vskip 0pt
			\centerline{\scriptsize {(34.90, 0.9535)}}\vskip -3pt
		\end{minipage}	
		\caption{Recovered frames for $Akiyo$ by different algorithms. The first, second and third rows are the restoration results of $1$th, $25$th and $50$th frames, respectively.}
		\label{Akiyo_SR20}
	\end{figure*}

	\begin{figure*}[htb]
		\centering
		\begin{minipage}{0.0915\linewidth}
			\caption*{Original}
			\vspace{-6pt}
			\footnotesize
			\centerline{\includegraphics[width=1.95cm]{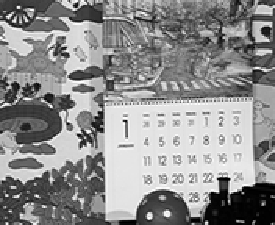}}\vskip 0pt
			\centerline{}\vskip -3pt
		\end{minipage}\hspace{2mm}
		\begin{minipage}{0.0915\linewidth}
			\caption*{Observed}
			\vspace{-5pt}
			\footnotesize
			\centerline{\includegraphics[width=1.95cm]{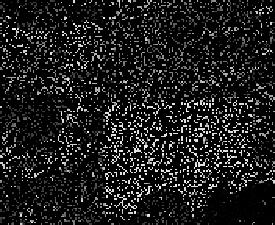}}\vskip 0pt
			\centerline{\scriptsize {(PSNR, SSIM)}}\vskip -3pt
		\end{minipage}\hspace{2mm}
		\begin{minipage}{0.0915\linewidth}
			\footnotesize
			\caption*{PSTNN}
			\vspace{-5pt}
			\centerline{\includegraphics[width=1.95cm]{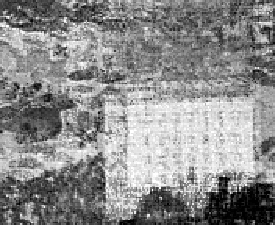}}\vskip 0pt
			\centerline{\scriptsize {{(16.56, 0.4897)}}}\vskip -3pt
		\end{minipage}\hspace{2mm}
		\begin{minipage}{0.0915\linewidth}
			\footnotesize
			\caption*{IRTNN}
			\vspace{-5pt}
			\centerline{\includegraphics[width=1.95cm]{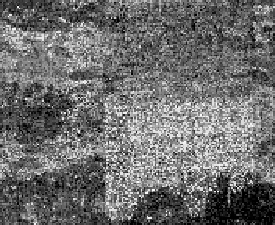}}\vskip 0pt
			\centerline{\scriptsize {(12.08, 0.2609)}}\vskip -3pt
		\end{minipage}\hspace{2mm}
		\begin{minipage}{0.0915\linewidth}
			\footnotesize
			\caption*{WTNN}
			\vspace{-5pt}
			\centerline{\includegraphics[width=1.95cm]{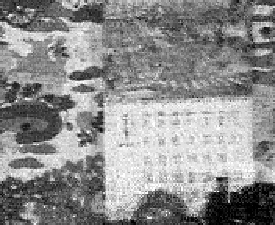}}\vskip 0pt
			\centerline{\scriptsize {(18.96, 0.5732)}}\vskip -3pt
		\end{minipage}\hspace{2mm}
		\begin{minipage}{0.0915\linewidth}
			\footnotesize
			\caption*{TNN}
			\vspace{-5pt}
			\centerline{\includegraphics[width=1.95cm]{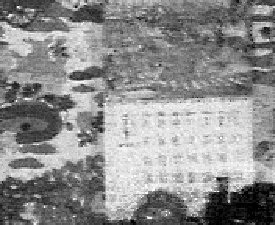}}\vskip 0pt
			\centerline{\scriptsize {(19.55, 0.6155)}}\vskip -3pt
		\end{minipage}\hspace{2mm}
		\begin{minipage}{0.0915\linewidth}
			\footnotesize
			\caption*{GTNN-HOP$_{0.6}$}
			\vspace{-5pt}
			\centerline{\includegraphics[width=1.95cm]{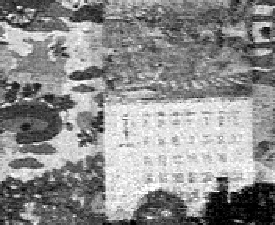}}\vskip 0pt
			\centerline{\scriptsize {(19.89, 0.6378)}}\vskip -3pt
		\end{minipage}\hspace{2mm}
		\begin{minipage}{0.0915\linewidth}
			\footnotesize
			\caption*{GTNN-HOP$_{0.3}$}
			\vspace{-5pt}
			\centerline{\includegraphics[width=1.95cm]{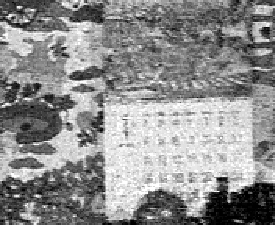}}\vskip 0pt
			\centerline{\scriptsize {(19.61, 0.6278)}}\vskip -3pt
		\end{minipage}\hspace{2mm}
		\begin{minipage}{0.0915\linewidth}
			\footnotesize
			\caption*{GTNN-HOC}
			\vspace{-5pt}
			\centerline{\includegraphics[width=1.95cm]{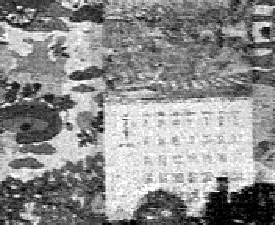}}\vskip 0pt
			\centerline{\scriptsize {(19.70, 0.6332)}}\vskip -3pt
		\end{minipage}	
		
		\begin{minipage}{0.0915\linewidth}
			\vspace{0pt}
			\footnotesize
			\centerline{\includegraphics[width=1.95cm]{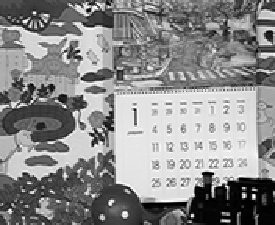}}\vskip 0pt
			\centerline{}\vskip -3pt
		\end{minipage}\hspace{2mm}
		\begin{minipage}{0.0915\linewidth}
			\vspace{1pt}
			\footnotesize
			\centerline{\includegraphics[width=1.95cm]{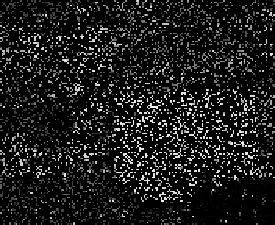}}\vskip 0pt
			\centerline{\scriptsize {(PSNR, SSIM)}}\vskip -3pt
		\end{minipage}\hspace{2mm}
		\begin{minipage}{0.0915\linewidth}
			\footnotesize
			\vspace{1pt}
			\centerline{\includegraphics[width=1.95cm]{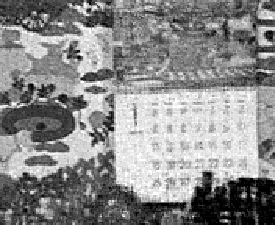}}\vskip 0pt
			\centerline{\scriptsize {{(21.42, 0.7129)}}}\vskip -3pt
		\end{minipage}\hspace{2mm}
		\begin{minipage}{0.0915\linewidth}
			\footnotesize
			\vspace{1pt}
			\centerline{\includegraphics[width=1.95cm]{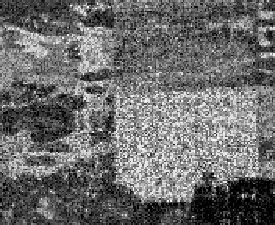}}\vskip 0pt
			\centerline{\scriptsize {(13.25, 0.3389)}}\vskip -3pt
		\end{minipage}\hspace{2mm}
		\begin{minipage}{0.0915\linewidth}
			\footnotesize
			\vspace{1pt}
			\centerline{\includegraphics[width=1.95cm]{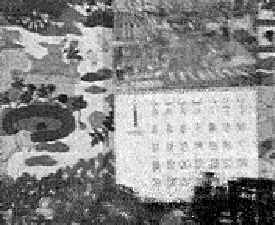}}\vskip 0pt
			\centerline{\scriptsize {(21.01, 0.6870)}}\vskip -3pt
		\end{minipage}\hspace{2mm}
		\begin{minipage}{0.0915\linewidth}
			\footnotesize
			\vspace{1pt}
			\centerline{\includegraphics[width=1.95cm]{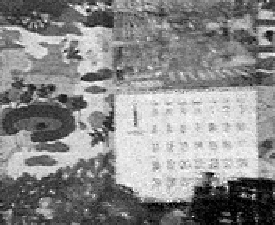}}\vskip 0pt
			\centerline{\scriptsize {(21.36, 0.7138)}}\vskip -3pt
		\end{minipage}\hspace{2mm}
		\begin{minipage}{0.0915\linewidth}
			\footnotesize
			\vspace{1pt}
			\centerline{\includegraphics[width=1.95cm]{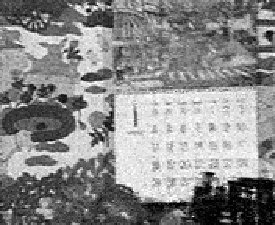}}\vskip 0pt
			\centerline{\scriptsize {(22.05, 0.7407)}}\vskip -3pt
		\end{minipage}\hspace{2mm}
		\begin{minipage}{0.0915\linewidth}
			\footnotesize
			\vspace{1pt}
			\centerline{\includegraphics[width=1.95cm]{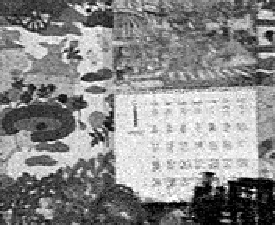}}\vskip 0pt
			\centerline{\scriptsize {(21.80, 0.7295)}}\vskip -3pt
		\end{minipage}\hspace{2mm}
		\begin{minipage}{0.0915\linewidth}
			\footnotesize
			\vspace{1pt}
			\centerline{\includegraphics[width=1.95cm]{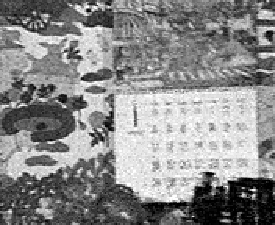}}\vskip 0pt
			\centerline{\scriptsize {(21.93, 0.7364)}}\vskip -3pt
		\end{minipage}	
		
		\begin{minipage}{0.0915\linewidth}
			\vspace{0pt}
			\footnotesize
			\centerline{\includegraphics[width=1.95cm]{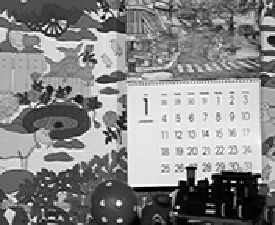}}\vskip 0pt
			\centerline{}\vskip -3pt
		\end{minipage}\hspace{2mm}
		\begin{minipage}{0.0915\linewidth}
			\vspace{1pt}
			\footnotesize
			\centerline{\includegraphics[width=1.95cm]{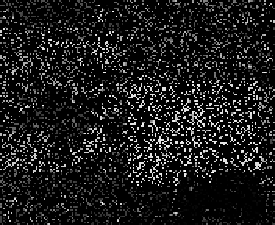}}\vskip 0pt
			\centerline{\scriptsize {(PSNR, SSIM)}}\vskip -3pt
		\end{minipage}\hspace{2mm}
		\begin{minipage}{0.0915\linewidth}
			\footnotesize
			\vspace{1pt}
			\centerline{\includegraphics[width=1.95cm]{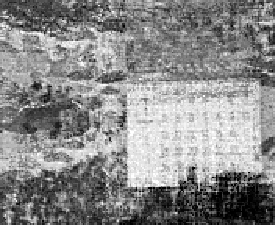}}\vskip 0pt
			\centerline{\scriptsize {{(16.42, 0.4903)}}}\vskip -3pt
		\end{minipage}\hspace{2mm}
		\begin{minipage}{0.0915\linewidth}
			\footnotesize
			\vspace{1pt}
			\centerline{\includegraphics[width=1.95cm]{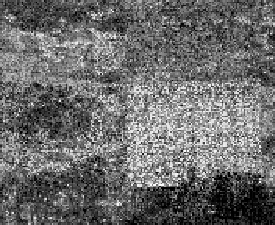}}\vskip 0pt
			\centerline{\scriptsize {(12.41, 0.2592)}}\vskip -3pt
		\end{minipage}\hspace{2mm}
		\begin{minipage}{0.0915\linewidth}
			\footnotesize
			\vspace{1pt}
			\centerline{\includegraphics[width=1.95cm]{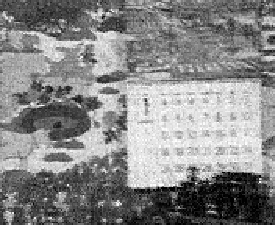}}\vskip 0pt
			\centerline{\scriptsize {(19.65, 0.6307)}}\vskip -3pt
		\end{minipage}\hspace{2mm}
		\begin{minipage}{0.0915\linewidth}
			\footnotesize
			\vspace{1pt}
			\centerline{\includegraphics[width=1.95cm]{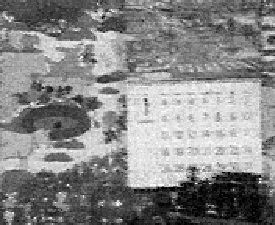}}\vskip 0pt
			\centerline{\scriptsize {(20.12, 0.6665)}}\vskip -3pt
		\end{minipage}\hspace{2mm}
		\begin{minipage}{0.0915\linewidth}
			\footnotesize
			\vspace{1pt}
			\centerline{\includegraphics[width=1.95cm]{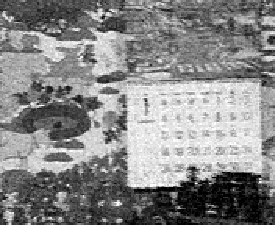}}\vskip 0pt
			\centerline{\scriptsize {(20.48, 0.6842)}}\vskip -3pt
		\end{minipage}\hspace{2mm}
		\begin{minipage}{0.0915\linewidth}
			\footnotesize
			\vspace{1pt}
			\centerline{\includegraphics[width=1.95cm]{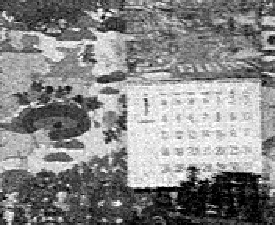}}\vskip 0pt
			\centerline{\scriptsize {(20.15, 0.6708)}}\vskip -3pt
		\end{minipage}\hspace{2mm}
		\begin{minipage}{0.0915\linewidth}
			\footnotesize
			\vspace{1pt}
			\centerline{\includegraphics[width=1.95cm]{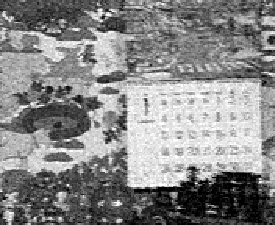}}\vskip 0pt
			\centerline{\scriptsize {(20.25, 0.6774)}}\vskip -3pt
		\end{minipage}	
		\caption{Recovered frames for $Mobile$ by different algorithms. The first, second and third rows are the restoration results of $1$th, $25$th and $50$th frames, respectively.}
		\label{Mobile_SR20}
	\end{figure*}

	\begin{figure}
		\centering
		\begin{minipage}{0.23\linewidth}
			\footnotesize
			\vspace{1pt}
			\centerline{\includegraphics[width=4.2cm]{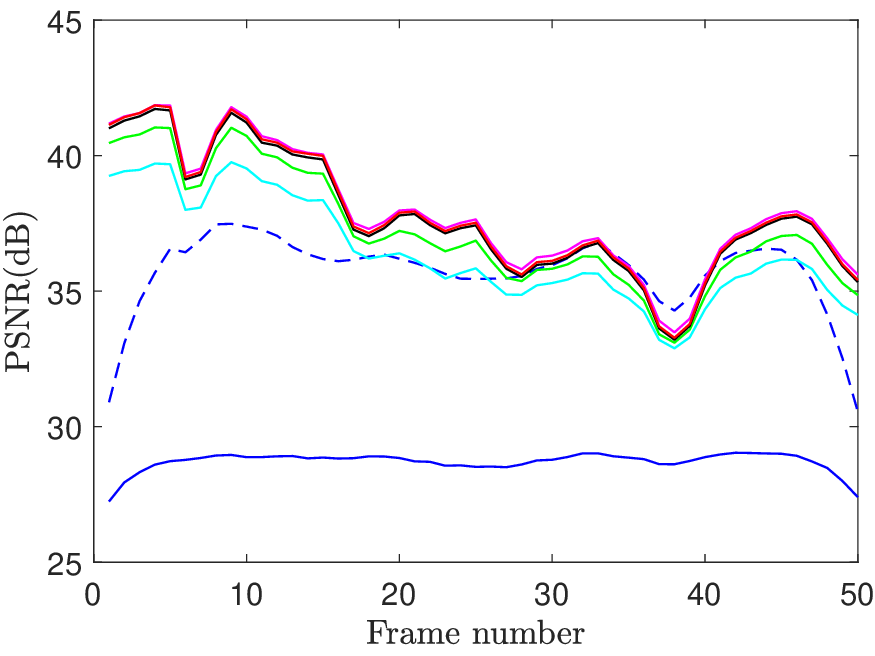}}\vskip 0pt
			\centerline{\scriptsize {(a). PSNR value of each frame for}}\vskip -1pt
			\centerline{\scriptsize { $Akiyo$ with ${\rm SR}=20\%$.}}\vskip -3pt
			\centerline{ }
		\end{minipage}\hspace{23mm}
		\begin{minipage}{0.23\linewidth}
			\footnotesize
			\vspace{1pt}
			\centerline{\includegraphics[width=4.2cm]{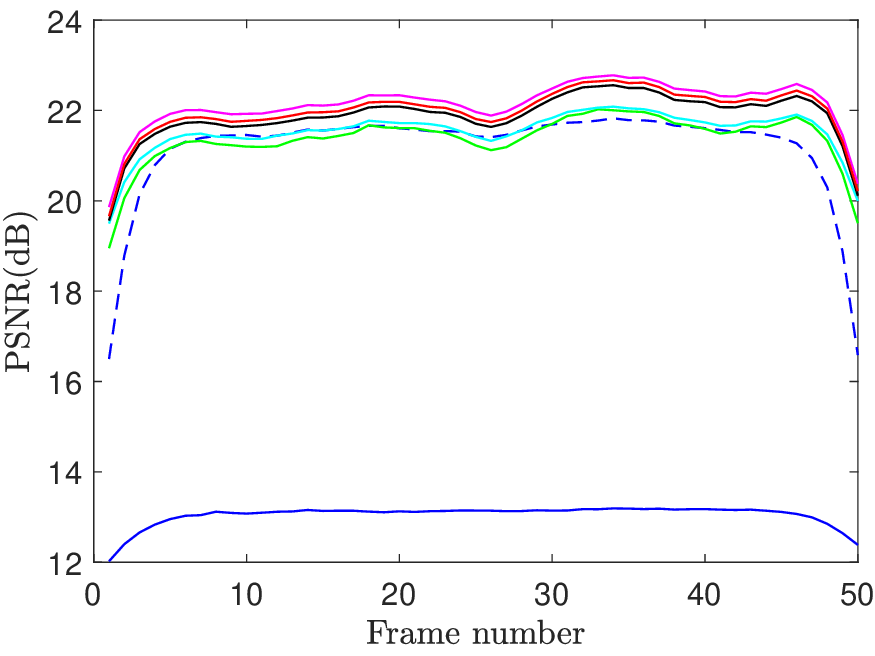}}\vskip 0pt
			\centerline{\scriptsize {(b). PSNR value of each frame for}}\vskip -1pt
			\centerline{\scriptsize { $Mobile$ with ${\rm SR}=20\%$.}}\vskip -3pt
			\centerline{ }
		\end{minipage}
		
		\begin{minipage}{0.23\linewidth}
			\footnotesize
			\vspace{1pt}
			\centerline{\includegraphics[width=4.2cm]{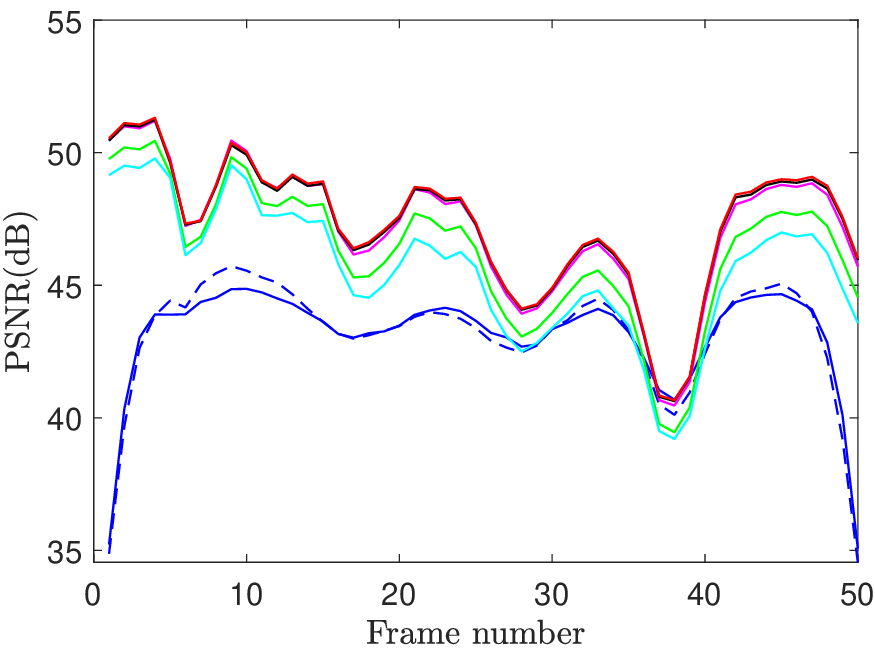}}\vskip 0pt
			\centerline{\scriptsize {(c). PSNR value of each frame for}}\vskip -1pt
			\centerline{\scriptsize { $Akiyo$ with ${\rm SR}=50\%$.}}\vskip -3pt
			\centerline{ }
		\end{minipage}\hspace{23mm}
		\begin{minipage}{0.23\linewidth}
			\footnotesize
			\vspace{1pt}
			\centerline{\includegraphics[width=4.2cm]{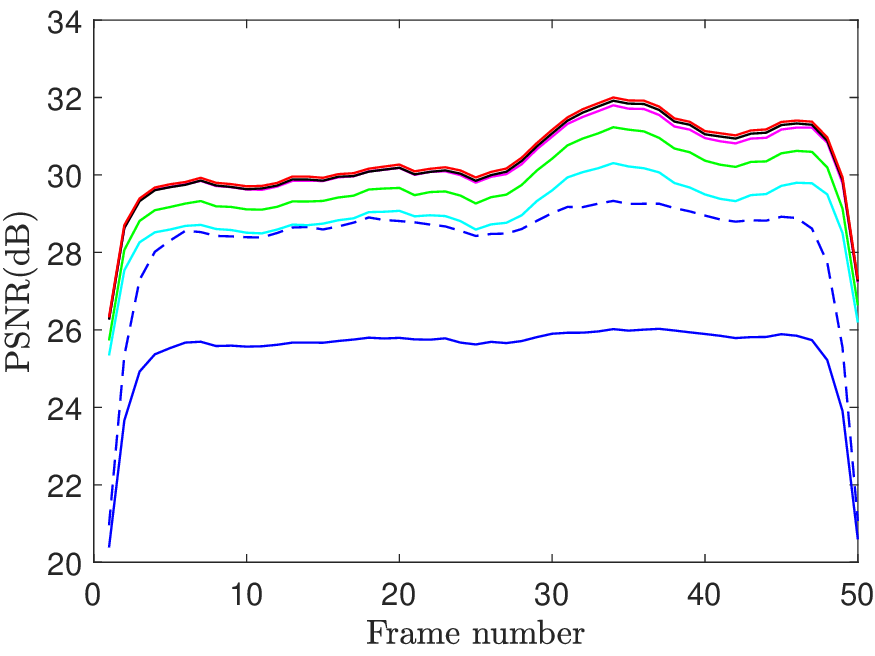}}\vskip 0pt
			\centerline{\scriptsize {(a). PSNR value of each frame for}}\vskip -1pt
			\centerline{\scriptsize { $Mobile$ with ${\rm SR}=50\%$.}}\vskip -3pt
			\centerline{ }
		\end{minipage}
	
		\begin{minipage}{0.45\linewidth}
			\footnotesize
			\vspace{1pt}
			\centerline{\includegraphics[width=8.0cm]{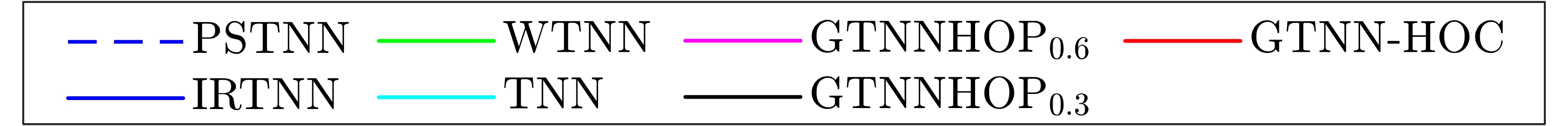}}\vskip 0pt
			\centerline{ }
		\end{minipage}
		\caption{Restoration results of each frame in terms of average PSNR value.}
		\label{PSNR_per_frame}
	\end{figure}
	
	\subsection{Video Restoration}
	
	The gray video sequences can be modeled as a $3$rd-order tensor and have a notable low-tubal-rank structure because of redundant information between frames. We test all algorithms on the YUV Video Sequences\footnote{\url{http://trace.eas.asu.edu/yuv/}}, and two videos, namely, $Akiyo$ and $Mobile$, are chosen. The frame sizes are $147\times 176$, and the first $50$ frames are used~\cite{KongHlpTNN2018}. Thus, the dimensions of each videos are $144\times 176\times 50$, and GTNN-HOP as well as GTNN-HOC are used to recover the incomplete video sequences. Fig.~\ref{Akiyo_SR20} shows three recovered frames ($1$th, $25$th and $50$th frames) of the $Akiyo$ with ${\rm SR}=20\%$, namely, $80\%$ randomly missing pixels. It is seen that our algorithms provide better recovery performance than the PSTNN, IRTNN, WTNN and TNN in terms of PSNR value, and GTNN-HOP$_{0.6}$ achieves the highest PSNR and SSIM values than the remaining approaches. To contrast the performance of algorithms on different frames, Fig.~\ref{PSNR_per_frame} (a) plots the average PSNR of all frames. We observe that the average PSNR values of our methods are higher than those of the competing techniques for most of the frames.
	Similarly, the restoration results of $Mobile$ are shown in Figs.~\ref{Mobile_SR20} and~\ref{PSNR_per_frame} (b). Again, the PSNR and SSIM values of the proposed approaches are higher than those of the competitors and GTNN-HOP$_{0.6}$ attains the best recovery performance in terms of PSNR and SSIM values. It is also illustrated in Fig.~\ref{PSNR_per_frame} (b) that our methods outperform the competing algorithms for all frames.
	
	On the other hand, the recovery performance of all methods under a higher SR, i.e., ${\rm SR} = 50\%$, is investigated. The restoration results of $Akiyo$ and $Mobile$ are shown in Figs.~\ref{PSNR_per_frame} (c) and~(d), respectively. We see that the developed algorithms are still superior to the competing techniques. 
	Furthermore, the running times of all methods are tabulated in Table~\ref{Video_runtime}. It is observed that TNN needs the least runtime while our approaches are faster than the remaining competitors.

	\begin{table}[htb]
		\caption{\small {Runtime of video restoration by different methods. The results are averages of $20$ independent runs.}}  
		\begin{center}
			\setlength{\tabcolsep}{3mm}{
				\begin{tabular}{ccccc}
					\hline
					\multirow{2}{*}{\begin{tabular}[c]{@{}c@{}}Method\end{tabular}} & \multicolumn {2} {c} {${\rm SR}=20\%$} &\multicolumn {2} {c} {${\rm SR}=50\%$} \\
					\cmidrule(lr){2-3} \cmidrule(lr){4-5}					
					& $Akiyo$& $Mobile$& $Akiyo$& $Mobile$\\
					\hline
					\multirow{1}{*}{\begin{tabular}[c]{@{}c@{}}PSTNN\cite{JiangTXPSTNN2020}\end{tabular}}
					& \multicolumn{1}{c}{25.53} &\multicolumn{1}{c}{22.63} & \multicolumn{1}{c}{24.35} &\multicolumn{1}{c}{{21.04}} \\
					\hline
					\multirow{1}{*}{\begin{tabular}[c]{@{}c@{}}IRTNN\cite{WangHIRTNN2022}\end{tabular}}
					& \multicolumn{1}{c}{532.6} &\multicolumn{1}{c}{678.0} & \multicolumn{1}{c}{146.4} &\multicolumn{1}{c}{{307.5}}\\
					\hline
					\multirow{1}{*}{\begin{tabular}[c]{@{}c@{}}WTNN\cite{MUYWTNN2020}\end{tabular}}
					& \multicolumn{1}{c}{9.499} &\multicolumn{1}{c}{9.163} & \multicolumn{1}{c}{9.647} &\multicolumn{1}{c}{{9.244}}\\
					\hline
					\multirow{1}{*}{\begin{tabular}[c]{@{}c@{}}TNN\cite{LuCTNN2018}\end{tabular}}
					& \multicolumn{1}{c}{\bf 7.600} &\multicolumn{1}{c}{\bf 7.086} & \multicolumn{1}{c}{\bf 7.417} &\multicolumn{1}{c}{{\bf 6.725}}\\
					\hline
					\multirow{1}{*}{\begin{tabular}[c]{@{}c@{}}GTNN-HOP$_{0.6}$\end{tabular}}
					& \multicolumn{1}{c}{9.016} &\multicolumn{1}{c}{9.157} & \multicolumn{1}{c}{8.299} &\multicolumn{1}{c}{{7.967}}\\
					\hline
					\multirow{1}{*}{\begin{tabular}[c]{@{}c@{}}GTNN-HOP$_{0.3}$\end{tabular}}
					& \multicolumn{1}{c}{9.366} &\multicolumn{1}{c}{8.818} & \multicolumn{1}{c}{8.255} &\multicolumn{1}{c}{{8.400}}\\
					\hline
					\multirow{1}{*}{\begin{tabular}[c]{@{}c@{}}GTNN-HOC\end{tabular}}
					& \multicolumn{1}{c}{9.114} &\multicolumn{1}{c}{8.799} & \multicolumn{1}{c}{8.281} &\multicolumn{1}{c}{{8.264}}\\
					\hline
			\end{tabular}}
			\vspace{-2em}
			\label{Video_runtime}
		\end{center}
	\end{table}

	\section{Conclusion}\label{Conclusion}
	In this paper, we have devised a framework to generate sparsity-inducing regularizers via half-quadratic optimization to alleviate the bias generated by the $\ell_1$-norm, and have solved the shortcoming that some nonconvex surrogates do not provide the closed-form expressions for their thresholding functions. To verify the effectiveness of these regularizers, we apply them to LRTC and propose the GTSVT operator. Furthermore, algorithms based on the ADMM  are developed. We analyze that the sequences generated by our algorithms are bounded, and prove that any limit point satisfies the KKT conditions. Extensive numerical examples based on synthetic and real-world datasets demonstrate that the recovery performance of the developed algorithms is superior to that of the competitors and our approaches need far less runtime than the IRTNN because the latter involves iterations to find its thresholding function.

\end{document}